\documentclass{article} 
\usepackage{iclr2027_conference,times}

\usepackage{amsmath,amsfonts,bm}

\def\eqref#1{equation~\ref{#1}}

\def\1{\bm{1}}

\DeclareMathAlphabet{\mathsfit}{\encodingdefault}{\sfdefault}{m}{sl}
\SetMathAlphabet{\mathsfit}{bold}{\encodingdefault}{\sfdefault}{bx}{n}

\usepackage{hyperref}
\usepackage{url}

\usepackage{pifont}
\usepackage{bbding}
\usepackage{xspace}
\usepackage{colortbl}
\usepackage{graphicx}
\usepackage{wrapfig}
\usepackage{amsmath,amsthm}
\usepackage{adjustbox}
\usepackage{caption}
\usepackage{zref-abspos}
\usepackage{tcolorbox}

\usepackage{enumitem}
\usepackage{amssymb}
\usepackage{algorithm}
\usepackage{algorithmic}
\usepackage{array}
\newcolumntype{C}{>{\centering\arraybackslash}X} 
\usepackage{tabularx}
\usepackage{booktabs}
\usepackage{diagbox}
\usepackage{multirow}
\usepackage{makecell}
\usepackage{minitoc}
\usepackage{etoc}
\usepackage{relsize}
\newcolumntype{Z}[1]{>{\centering\arraybackslash\larger[1]}p{#1}}

\newcommand{{\method}}{\textsc{Seg-OPD}\xspace}

\DeclareUnicodeCharacter{FF1F}{?} 

\title{Learning to Revise Reasoning with Segment-wise On-Policy Distillation}

\author{%
\textbf{Yuxiang Zhang\textsuperscript{1},
Ding Cao\textsuperscript{1,2},
Shuting Cui\textsuperscript{1}} \\
\textbf{Lei Wang\textsuperscript{3},
Weijieying Ren\textsuperscript{4},
Tianxiang Zhao\textsuperscript{1}}\thanks{Corresponding author.} \\[0.5em]
{\small
\textsuperscript{1}\,The Hong Kong University of Science and Technology (Guangzhou)} \\
{\small
\textsuperscript{2}\,University of Science and Technology of China} \\
{\small
\textsuperscript{3}\,Alibaba Group} \\
{\small
\textsuperscript{4}\,Stanford University} \\[0.5em]
{\fontsize{8}{10}\selectfont\texttt{\{yzhang800,scui847\}@connect.hkust-gz.edu.cn, caoding@mail.ustc.edu.cn}} \\
{\fontsize{8}{10}\selectfont\texttt{lw.lld3b8@alibaba-inc.com, wjyren@stanford.edu, tianxiangz@hkust-gz.edu.cn}}
}

\definecolor{gold}{RGB}{184,134,11}
\newcommand{\secondbest}[1]{\textcolor{gold}{\underline{#1}}}

\iclrfinalcopy 
\begin{document}

\maketitle
\lhead{Preprint}

\begin{abstract}
On-policy distillation (OPD) improves large language model reasoning by training students on their own rollouts with dense token-wise supervision from the teacher.
However, token-wise OPD does not explicitly provide a coherent alternative reasoning step showing how the student's step could be revised to improve subsequent reasoning.
Furthermore, this paradigm can become less effective when the student produces a degenerate reasoning prefix, as subsequent teacher supervision remains conditioned on that prefix and may reinforce poor reasoning patterns.
In this work, we focus on learning reasoning revision with segment-wise OPD to rework intermediate reasoning steps and better support subsequent reasoning.
Through controlled reasoning interventions, we find that replacing student segments with teacher redrafts improves subsequent reasoning accuracy.
Therefore, we address the problem of turning teacher redrafts into explicit supervision for learning to revise reasoning.
We propose Segment-wise On-Policy Distillation (\method), which selects student segments based on an uncertainty metric and obtains corresponding teacher redrafts.
\method trains the student to prefer teacher redrafts over their paired student segments while retaining dense token-wise OPD supervision.
Extensive experiments on mathematical reasoning and competitive programming tasks show that \method-trained students achieve higher revision success rates than baselines.
\method consistently outperforms the compared state-of-the-art baselines in reasoning accuracy with an average relative improvement of 5.22\% across diverse models and tasks.
Code is available at \href{https://anonymous.4open.science/r/Seg-OPD}{https://anonymous.4open.science/r/Seg-OPD}.

\vspace{-2.5em}
\end{abstract}

\begin{figure*}[b!]
    \centering
    \includegraphics[width=\textwidth]{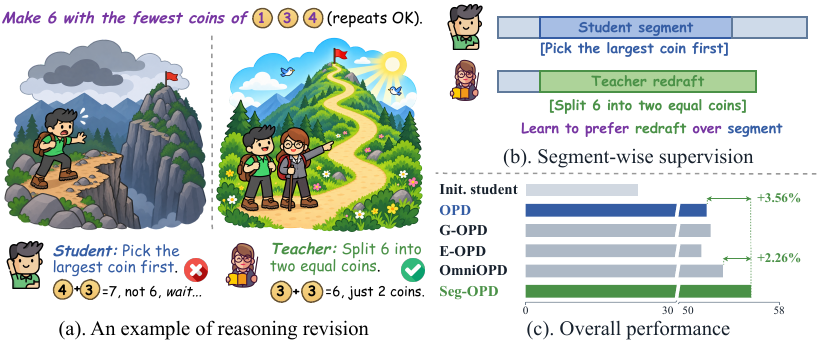}
    \caption{\textbf{(a). An example of reasoning revision.} The student's largest-coin-first attempt overshoots the target, whereas the teacher redraft reaches it with two equal-value coins. \textbf{(b). Segment-wise supervision.} The paired student segment and teacher redraft illustrate explicit supervision for learning reasoning revision. \textbf{(c). Overall Performance.} Reasoning accuracy (\%) of \method and baselines for a Qwen3-14B student on AIME25. 
    }
    \label{fig:illustration}
\end{figure*}

\section{Introduction}


On-Policy Distillation (OPD) has emerged as an effective paradigm for LLM post-training~\citep{agarwal2024policy}.
OPD has the student generate its own reasoning trajectories and prompts a stronger teacher to provide feedback.
By integrating the dense token-wise supervision of Supervised Fine-Tuning (SFT) into the on-policy exploration of Reinforcement Learning (RL), OPD mitigates both the distribution shift introduced by imitating offline demonstrations in SFT~\citep{ouyang2022training} and the sparse reward signals of outcome-based RL~\citep{shao2024deepseekmath}.


Despite these strengths, OPD suffers from two inherent limitations.
\emph{First,} because vanilla OPD provides only local corrections through token-wise supervision, the student is never explicitly shown the coherent reasoning step that spans multiple tokens.
In such cases, local corrections do not directly show the student how to revise the reasoning step as a whole and redirect subsequent reasoning.
\emph{Second,} due to LLMs' autoregressive nature, their subsequent reasoning is conditioned on the prefix. 
Therefore, the existing degenerate prefix from the student rollout can be harmful and thus degrade the teacher's guidance, leading to suboptimal student learning.
Recent studies show that on degenerate prefixes, teacher feedback can favor unproductive continuations rather than penalize them, potentially reinforcing patterns such as repetition loops~\citep{luo2026demystifying, fu2026revisiting}.

These limitations motivate us to improve vanilla OPD by replacing unproductive reasoning steps and redirecting the student's subsequent reasoning.
More precisely, given a prefix and a student reasoning segment, the teacher should provide a coherent redraft that revises the corresponding segment while preserving the prefix.
Such a redraft retains the on-policy context that exposes the student's current unproductive reasoning mode, providing an alternative reasoning transition that can redirect subsequent reasoning. 
We use \textbf{reasoning revision} to describe reworking an intermediate reasoning step to better support subsequent reasoning, and illustrate it in Fig.~\ref{fig:illustration}.
In Fig.~\ref{fig:illustration}(a), the teacher redraft offers an equal-split strategy instead of the student's largest-coin-first approach.
Fig.~\ref{fig:illustration}(b) illustrates the use of teacher redrafts as explicit supervision for learning reasoning revision.

Though promising, implementing reasoning revision learning faces two challenges. 
\emph{First,} teacher redrafting is costly, so how can we fully utilize the limited budget to select student segments worth revising and obtain suitable teacher redrafts?
\emph{Second,} how can the learning objective exploit the correspondence between student segments and teacher redrafts while preserving the dense token-wise supervision that makes OPD effective?
To address these challenges, we conduct a preliminary test-time analysis in Sec.~\ref{sec:preliminary} to inform our method's design. 
We reveal that replacing selected student segments with teacher redrafts improves subsequent reasoning accuracy for both the initial and OPD-trained student models. 
Furthermore, we find that student reasoning pivots associated with high uncertainty provide more useful revision targets than random anchors.

Building upon these insights, we propose a novel \underline{Seg}ment-wise \underline{O}n-\underline{P}olicy \underline{D}istillation method (\method).
To better utilize the limited budget, \method uses an entropy-based uncertainty metric to guide student segment selection and obtains corresponding teacher redrafts, providing targeted revision supervision.
\method employs a revision learning objective that encourages the student to prefer teacher redrafts over the paired student segments through a segment-wise preference loss, while retaining dense token-wise OPD supervision.
As Fig.~\ref{fig:illustration}(c) shows, \method outperforms the strongest baseline by 2.26\% in reasoning accuracy with Qwen3-14B on AIME25.

We summarize our contributions as follows: 
\begin{itemize}[leftmargin=*,topsep=0pt]
    \item We formulate the problem of reasoning revision in OPD as learning from teacher redrafts of students' own intermediate reasoning.
    Through controlled trajectory interventions, we show that replacing student segments with teacher redrafts substantially improves downstream reasoning, establishing teacher redrafts as useful explicit revision supervision;

    \item We introduce \method, which enables students to learn reasoning revision from revision units that pair their own reasoning segments with corresponding teacher redrafts.
    To achieve this, \method trains students to prefer the redrafts over their paired student segments while retaining dense token-wise OPD feedback;

    \item We conduct experiments with diverse student models across a wide range of math and code benchmarks.
    \method outperforms the compared state-of-the-art OPD methods in reasoning accuracy, achieving a 5.22\% average relative improvement over competitors.
    Beyond accuracy, \method also achieves higher reasoning revision success rates and fewer reasoning loops after training. 
\end{itemize}


\section{Related Work}\label{app:related-work}
\subsection{On-Policy Distillation}
On-policy Distillation (OPD) trains students on their own rollouts using teacher supervision~\citep{agarwal2024policy}.
Recently, OPD has been extended to multi-teacher, multi-task post-training~\citep{xiao2026mimo,xu2026deepseek} and on-policy self-distillation~\citep{zhao2026self,zhao2026rosd}.
Although OPD's effectiveness primarily depends on learning the teacher's reasoning behavior~\citep{li2026rethinking,luo2026demystifying}, how these patterns transfer remains underexplored~\citep{fu2026revisiting,yang2026learning,cai2026learning}.
By explicitly guiding the student to prefer the teacher's redraft over its own segment, \method emphasizes this reasoning behavior during training.

\subsection{Learning from Reasoning Interventions}

As LLMs reason autoregressively, early errors can misdirect subsequent steps, motivating interventions~\citep{lai2024step,xu2026pass}.
Prior work trains reward models with step-level labels~\citep{lightman2024lets,wang2024math}, compares reasoning steps for preference learning~\citep{lai2024step,zhang2024chain,yang2025supercorrect}, or introduces teacher guidance during on-policy training~\citep{xu2026pass,huang2026learning,zhao2026rosd}.
Unlike offline step-pairing~\citep{lai2024step,zhang2024chain} or interleaving teacher generations within a rollout~\citep{xu2026pass}, \method pairs student segments with teacher redrafts, training the student to prefer these redrafts over its own segments while retaining token-wise OPD on the remaining tokens.

\section{Preliminary Analysis}\label{sec:preliminary}

In this section, we present the preliminaries on reasoning revision in on-policy distillation and empirical cases that motivate the methodology of our \method.

\begin{figure*}[t]
    \centering
    \includegraphics[width=\textwidth]{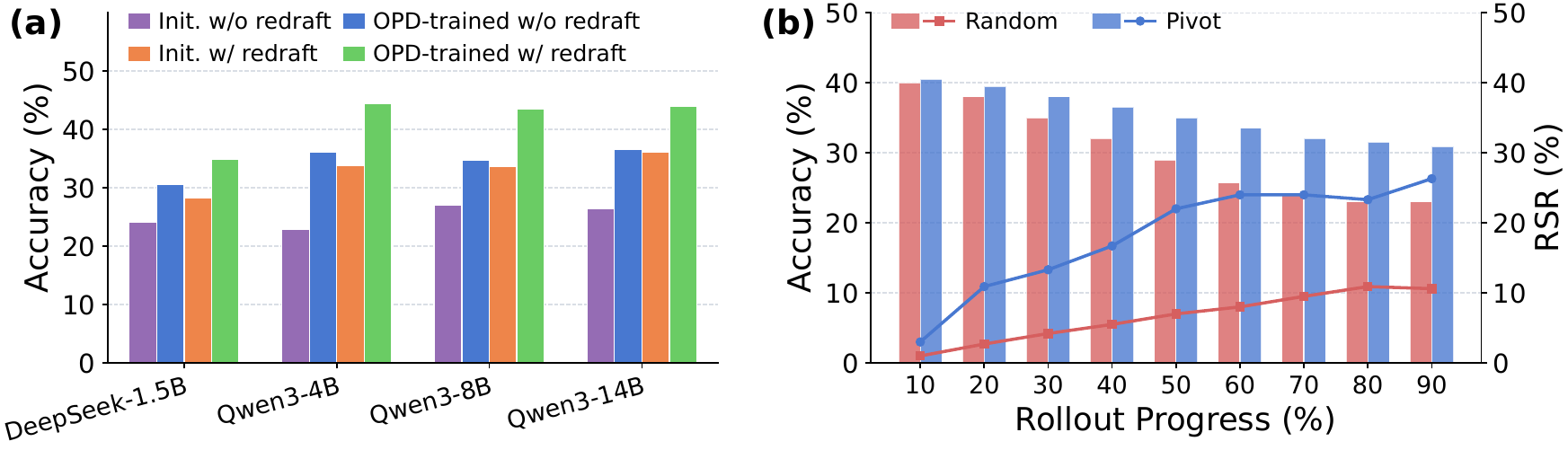}
    \caption{\textbf{Effectiveness of teacher redrafts and segment selection.}
    \textbf{(a).} Reasoning accuracy of initial/OPD-trained students with/without teacher redrafts on AIME24. 
    \textbf{(b).} After replacing segments selected with random/pivot anchors with teacher redrafts, the reasoning accuracy (bars, left axis) and revision success rate (lines, right axis) of Qwen3-4B on AIME26. 
    }
    \vspace{-1em}
    \label{fig:1}
\end{figure*}

\subsection{Notations}\label{sec:notation}

Let $x = (x_1, x_2, \ldots, x_n)$ be an input prompt and $y = (y_1, y_2, \ldots, y_m)$ a response. 
We denote the prefix up to step $t$ by $q_t=(x, y_{<t})$, where $y_{<t} \triangleq (y_1, y_2, \ldots, y_{t-1})$.
OPD setting generally involves two LLMs: a trainable student $\pi_\theta$ and a fixed teacher $\pi_*$.

\subsection{On-Policy Distillation}\label{sec:opd}

By exploring its own action space and receiving dense feedback from the teacher distribution $\pi_*$, OPD enables the student $\pi_{\theta}$ to learn from its own rollouts as:
\begin{equation}\label{eq:opd-objective}
\mathcal{J}_{\mathrm{OPD}}(\theta)=\mathbb{E}_{x\sim\mathcal{D}}\left[D_{\mathrm{KL}}\!\left(\pi_\theta(\cdot\mid x)\,\|\,\pi_*(\cdot\mid x)\right)\right].
\end{equation}
For $x\sim\mathcal{D}$ and $y\sim\pi_\theta(\cdot\mid x)$, the gradient of Eq.~(\ref{eq:opd-objective}) can be written as:

\begin{equation}\label{eq:opd-gradient}
\nabla_\theta\mathcal{J}_{\mathrm{OPD}}(\theta)=\mathbb{E}_{x,y}\left[\sum_{t=1}^{m}\left(\sum_{t'=t}^{m}\left[\log\pi_\theta(y_{t'}\mid q_{t'})-\log\pi_*(y_{t'}\mid q_{t'})\right]\right)\nabla_\theta\log\pi_\theta(y_t\mid q_t)\right].
\end{equation}

In practice, OPD methods use a discount factor of zero to focus on next-token optimization and thus approximate the gradient in Eq.~\ref{eq:opd-gradient} as:

\begin{equation}\label{eq:token-opd-gradient}
\nabla_\theta\mathcal{J}_{\mathrm{OPD}}(\theta)\approx\mathbb{E}_{x,y}\left[\sum_{t=1}^{m}\left[\log\pi_\theta(y_t\mid q_t)-\log\pi_*(y_t\mid q_t)\right]\nabla_\theta\log\pi_\theta(y_t\mid q_t)\right].
\end{equation}

The approximation in Eq.~\ref{eq:token-opd-gradient} is commonly called token-wise OPD.
Although computationally efficient, token-wise OPD provides feedback only for individual tokens, without explicitly presenting a coherent alternative reasoning step as a revision target.
We therefore consider teacher redrafts as explicit revision targets for student segments.
The segment-wise preference objective of \method uses conditional joint probabilities to compare each student segment with its teacher redraft.
To assess the value of these revision targets, we first examine whether replacing a student segment with a teacher redraft at test time improves subsequent reasoning accuracy.

\subsection{Understanding Reasoning Revision in OPD}\label{sec:casestudy}

\noindent\textbf{Teacher Redrafts Can Redirect Subsequent Reasoning.}
As LLMs generate responses autoregressively, each reasoning token is conditioned on the preceding context.
This dependency suggests that replacing an incorrect intermediate step with a better redraft can redirect subsequent reasoning.
To validate this, we first manually locate the wrong reasoning segment $s$ in the student rollout, and then assign the same prefix $q$ to a teacher model to generate a redraft $r$.
For comparison, we prompt the student to continue the subsequent reasoning from $(q,s)$ and $(q,r)$, respectively.
As shown in Fig.~\ref{fig:1}(a), replacing $s$ with $r$ increases reasoning accuracy for both initial and OPD-trained student models, indicating that teacher redrafts can provide direct and effective reasoning revisions.

\noindent\textbf{Reasoning Pivots Provide Better Revision Targets.}
To extend these benefits to on-policy training, we need an efficient way to select informative segments worth revising without relying on manual review.
Prior work shows that high-entropy tokens indicate potential turning points in reasoning~\citep{wang2025beyond,cheng2026reasoning}.
Therefore, we select these pivots as anchors and expand them in both directions to period-marked sentence boundaries, forming a student segment for teacher redrafting.
To examine whether reasoning pivots provide better revision targets, we compare reasoning units centered on \textbf{(i) random anchors} and \textbf{(ii) pivot anchors} around the same rollout positions.
Beyond reasoning accuracy, we also report the revision success rate (RSR), defined as the proportion of student revision attempts in the complete trajectory that correct an earlier error with a substantively different alternative. 
For RSR evaluation, we call GPT-5.6 to extract reasoning steps from rollouts and assess whether the attempts meet this criterion.
Appendix~\ref{app:rsr-evaluation} provides more details.
As shown in Fig.~\ref{fig:1}(b), pivot anchors yield higher reasoning accuracy and higher student revision success rates than random anchors at matched rollout positions.


\noindent\textbf{Summary.} 
These results show that \textbf{(i)} replacing selected student segments with teacher redrafts can improve the student's subsequent reasoning accuracy; and \textbf{(ii)} at matched rollout positions, pivot anchors yield higher reasoning accuracy than random anchors and are accompanied by higher RSR.
The analysis further motivates a novel segment-wise OPD paradigm in which the student directly learns reasoning revisions from teacher redrafts on selected segments.

\begin{figure*}[t]
    \centering
    \includegraphics[width=\textwidth]{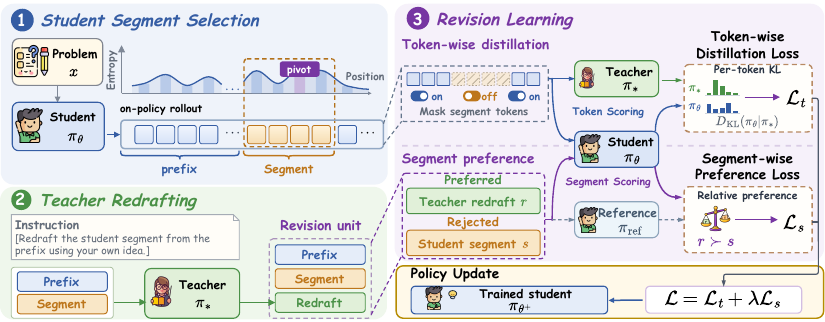}
    \caption{\textbf{Framework of \method.}
    \method enables students to learn reasoning revision by pairing student segments with teacher redrafts and using preference learning to favor these alternative reasoning steps over the corresponding student segments. Simultaneously, \method retains dense token-wise supervision outside the selected segments.
    }
    \label{fig:framework}
\end{figure*}

\section{Methodology}\label{sec:5}


Our preliminary analysis shows that teacher redraft can redirect students' subsequent reasoning, and reasoning pivots provide better revision targets. 
These findings leave two open questions for learning reasoning revision:
\textbf{(i)} How to select student segments and construct corresponding teacher redrafts for learning reasoning revision? 
\textbf{(ii)} How to derive the supervision signal from the revision unit to guide the student in learning reasoning revision?

To address these challenges, we propose \method, and illustrate its framework in Fig.~\ref{fig:framework}.
As Fig.~\ref{fig:framework} shows, \method first selects student segments around reasoning pivots in student rollouts.
For each selected segment, \method prompts the teacher to propose an alternative reasoning step under the same prefix, forming a revision unit from the prefix, student segment, and teacher redraft.
Consequently, \method trains the student by optimizing a segment-wise preference learning objective on these revision units while retaining token-wise OPD on the remaining valid rollout tokens.
Next, we present each component in detail.



\subsection{Revision Unit Construction}

\noindent\textbf{Segment Selection.}
Motivated by the analysis in Sec.~\ref{sec:casestudy}, we find that student segments constructed using reasoning pivots achieve higher revision success rates and higher reasoning accuracy after redrafting than those constructed using random anchors at the same rollout positions.
Since rollouts ending in incorrect results are more likely to contain reasoning steps that can benefit from revision, \method restricts candidate segment construction to these rollouts.

For each selected rollout $y$, we design a hierarchical uncertainty metric based on two considerations:
\textbf{(i)} a pivot anchor should exhibit high predictive uncertainty under the current student; and \textbf{(ii)} the complete reasoning segment containing the anchor should exhibit high aggregated uncertainty, rather than being selected because of an isolated high-entropy token.
We first quantify uncertainty at each position using the predictive entropy recorded during rollout generation:
\begin{equation}\label{eq:pivot-entropy}
H_t = -\sum_{v\in\mathcal{V}}\pi_\theta(v\mid q_t)\log \pi_\theta(v\mid q_t),
\end{equation}
where $\mathcal{V}$ denotes the token vocabulary. 
Then \method selects the positions with the largest entropy values as pivot anchors and denotes the resulting set as $T$.
For each pivot anchor $t_i \in T$, \method expands it in both directions to the surrounding sentence boundaries marked by periods, yielding a candidate segment $s_i=y_{s_i^{b}:s_i^{e}}$, where $s_i^{b}$ and $s_i^{e}$ index its first and last tokens.
Additional analysis on boundary construction is provided in Appendix~\ref{app:segment-selection-analysis}.
We score each candidate by its mean predictive entropy penalized by the entropy dispersion within the segment:
\begin{equation}\label{eq:segment-uncertainty}
U(s_i)
=
\frac{1}{|s_i|}
\sum_{t=s_i^{b}}^{s_i^{e}} H_t
-
\delta_i,\quad |\mathcal{C}_K(y)|\leq K,
\end{equation}
where $\delta_i$ denotes the standard deviation of the token-entropy values within $s_i$, and $|s_i|=s_i^{e}-s_i^{b}+1$ is its token count.
Then, we greedily select segments in descending score order to form the set $\mathcal{C}_K(y)$, with at most $K$ segments per rollout.

\noindent\textbf{Teacher Redrafting.}
To provide a revision as a local alternative to the selected student segment for preference learning, \method constructs a teacher redraft by revising the corresponding student reasoning step.
We obtain this alternative from one full teacher reasoning, allowing the teacher to develop its reasoning toward the solution.
To achieve this, \method first provides the teacher model with both the prefix $q$ and the student segment $s$.
The teacher is explicitly prompted to rewrite $s$ starting after $q$ and continue reasoning; 
\method then extracts the leading sentence as the teacher redraft $r$ using the same boundary rule as for student segments, provided that the full draft passes a task-specific correctness check.
The corresponding teacher redraft $r$, together with $q$ and $s$, forms a revision unit $p=(q,s,r)$.
Appendix~\ref{app:segment} provides more details on revision unit construction.


\subsection{Reasoning Revision Learning}
For each revision unit $p_i=(q_i,s_i,r_i)$, \method trains the student to prefer $r_i$ over $s_i$ by optimizing a segment-wise preference objective while retaining token-wise OPD outside $s_i$.

\noindent\textbf{Segment-Wise Preference Objective.}
We start with a preference learning objective similar to that used in prior work~\citep{rafailov2023direct}.
Specifically, we use $r\succ s$ as the training preference relation between each teacher redraft and its paired student segment under the same prefix.
Let $\pi_{\mathrm{ref}}$ be the base reference policy of the student.
Following~\citep{lu-etal-2024-eliminating}, we normalize each segment's log policy ratio by its token length and rescale the difference by the mean segment length, defining the revision margin as:

\begin{equation}\label{eq:revision-margin}
m_{\pi_\theta}(p)
=
\frac{|s|+|r|}{2}
\left[
\frac{1}{|r|}
\log\frac{\pi_\theta(r\mid q)}{\pi_{\mathrm{ref}}(r\mid q)}
-
\frac{1}{|s|}
\log\frac{\pi_\theta(s\mid q)}{\pi_{\mathrm{ref}}(s\mid q)}
\right].
\end{equation}
Here, $\log\pi_\theta(r\mid q)$ and $\log\pi_\theta(s\mid q)$ are the logarithms of the segments' conditional joint probabilities, obtained by summing the token log probabilities over $r$ and $s$, respectively.
A positive revision margin in Eq.~(\ref{eq:revision-margin}) means that the student's length-normalized log-probability gap between $r$ and $s$ exceeds that of the reference policy.
Using the Bradley--Terry form~\citep{bradley1952rank}, we model the preference $r\succ s$ with probability $\sigma(\beta m_{\pi_\theta}(p))$.
Let $\{p_i=(q_i,s_i,r_i)\}_{i=1}^{N}$ denote the $N$ revision units in the current training batch.
We take the negative log-likelihood of this probability and average over these units, which gives the segment-wise loss:
\begin{equation}\label{eq:segment-loss}
\mathcal{L}_s(\theta)
=
-\frac{1}{N}
\sum_{i=1}^{N}
\log\sigma\!\left(
\beta m_{\pi_\theta}(p_i)
\right).
\end{equation}
Here, $\sigma$ is the logistic sigmoid and $\beta>0$ controls the margin scale.
The loss assigns larger penalties to units with small or negative revision margins and decreases as the margin grows.


\noindent\textbf{Token-wise Distillation.}
To help stabilize training~\citep{grattafiori2024llama}, \method retains token-wise OPD alongside the segment-wise preference objective.
The token-wise loss provides dense teacher supervision on the remaining valid rollout tokens.
Specifically, for a sampled rollout $y=(y_1,\ldots,y_m)$, let $M_t\in\{0,1\}$ denote the token mask.
We set $M_t=0$ when $y_t$ lies in any student segment used in a revision unit and $M_t=1$ otherwise.
The mask only determines which positions are used in the token-wise loss and does not change the sampled rollout.
Following the token-wise OPD formulation in Eq.~(\ref{eq:token-opd-gradient}), the token-wise loss with segment masking is:

\begin{equation}\label{eq:token-loss}
\mathcal{L}_t(\theta)
=
\mathbb{E}_{x\sim\mathcal{D},\,y\sim\pi_\theta(\cdot\mid x)}
\left[
\sum_{t=1}^{m}
M_t
D_{\mathrm{KL}}\!\left(
\pi_\theta(\cdot\mid q_t)
\,\|\,
\pi_*(\cdot\mid q_t)
\right)
\right].
\end{equation}

\begin{table*}[!t]
    \centering
    \small
    \setlength{\tabcolsep}{3pt}
    \renewcommand{\arraystretch}{1.05}
    \caption{Main results on mathematical reasoning and competitive programming. We report the mean Pass@1 (\%) across diverse student models and datasets. \textbf{Bold} and \secondbest{underline} denote the best and second-best results per evaluation setting.}
    \label{tab:2}
    \resizebox{\textwidth}{!}{%
    \begin{tabular*}{\textwidth}{@{\extracolsep{\fill}}lcccccccc@{}}
        \toprule
        \textbf{Method} & \textbf{Codeforces} & \textbf{TACO} & \textbf{HumanEval+}
        & \textbf{AIME24} & \textbf{AIME25} & \textbf{AIME26} & \textbf{MATH500} & \textbf{MINERVA} \\
        \midrule
        \rowcolor{gray!12}
        \multicolumn{9}{c}{\textit{Student: DeepSeek-1.5B}} \\
        \midrule
        OPD           & 33.30 & 28.30 & 73.25 & 43.09 & 36.67 & \secondbest{33.33} & 86.45 & 37.22 \\
        E-OPD          & \secondbest{34.90} & 29.30 & 78.84 & 43.31 & 35.83 & 30.83 & 86.05 & 37.59 \\
        G-OPD & 33.75 & \secondbest{30.06} & 78.67 & \secondbest{50.00} & \secondbest{39.17} & 31.67 & \secondbest{86.85} & \secondbest{38.69} \\
        OmniOPD       & 33.90 & 29.70 & \secondbest{79.39} & 42.50 & 32.50 & 28.33 & 86.05 & 37.04 \\
        \rowcolor{blue!8}
        \method       & \textbf{36.30} & \textbf{31.30} & \textbf{81.20} & \textbf{51.33} & \textbf{40.19} & \textbf{35.25} & \textbf{89.37} & \textbf{39.36} \\
        \midrule

        \rowcolor{gray!12}
        \multicolumn{9}{c}{\textit{Student: Qwen3-4B}} \\
        \midrule
        OPD & 42.34 & \secondbest{43.32} & 82.30 & 63.15 & 61.07 & 49.69 & \textbf{90.71} & 45.89 \\
        E-OPD & \secondbest{43.90} & 40.97 & \secondbest{84.67} & 62.50 & \secondbest{61.61} & 51.26 & 90.01 & 46.38 \\
        G-OPD & 43.50 & 40.58 & 83.69 & 62.32 & 61.23 & \secondbest{52.17} & \secondbest{90.05} & \secondbest{47.57} \\
        OmniOPD & 40.04 & 38.68 & 82.62 & \secondbest{63.50} & 59.56 & 50.49 & 88.73 & 44.85 \\
        \rowcolor{blue!8}
        \method & \textbf{45.89} & \textbf{43.65} & \textbf{85.89} & \textbf{64.50} & \textbf{63.00} & \textbf{52.53} & 89.10 & \textbf{48.85} \\
        \midrule

        \rowcolor{gray!12}
        \multicolumn{9}{c}{\textit{Student: Qwen3-8B}} \\
        \midrule
        OPD & 46.41 & 45.07 & 76.59 & 59.17 & 55.60 & 44.32 & 87.85 & 43.01 \\
        E-OPD & 45.75 & 44.35 & 78.54 & \textbf{63.68} & \secondbest{57.99} & 42.64 & \secondbest{90.31} & \secondbest{44.85} \\
        G-OPD & 46.01 & 41.89 & \secondbest{80.27} & 60.83 & 52.50 & \secondbest{46.79} & \textbf{90.40} & 44.60 \\
        OmniOPD & \secondbest{46.86} & \textbf{45.46} & 78.51 & 55.38 & 57.08 & 46.60 & 88.30 & 43.66 \\
        \rowcolor{blue!8}
        \method & \textbf{48.86} & \secondbest{45.35} & \textbf{82.37} & \secondbest{63.37} & \textbf{59.67} & \textbf{47.19} & 88.36 & \textbf{45.91} \\
        \midrule

        \rowcolor{gray!12}
        \multicolumn{9}{c}{\textit{Student: Qwen3-14B}} \\
        \midrule
        OPD & 37.35 & 45.14 & 81.06 & 60.67 & 52.17 & 46.67 & 87.50 & 45.60 \\
        E-OPD & 36.17 & 45.81 & 81.59 & \secondbest{61.17} & 51.75 & \textbf{49.34} & 87.84 & 44.12 \\
        G-OPD & 36.59 & 45.71 & 82.07 & 60.83 & 52.50 & 48.97 & 87.40 & \textbf{47.22} \\
        OmniOPD & \secondbest{39.88} & \textbf{48.12} & \secondbest{82.69} & 59.97 & \secondbest{53.47} & 45.76 & \textbf{88.97} & 45.23 \\
        \rowcolor{blue!8}
        \method & \textbf{41.59} & \secondbest{47.68} & \textbf{83.37} & \textbf{63.33} & \textbf{55.73} & \secondbest{49.17} & \secondbest{88.40} & \secondbest{46.00} \\
        \bottomrule
    \end{tabular*}%
    }
\end{table*}


Here, we present the final optimization function of \method, based on the analysis above.
We jointly optimize the token-wise distillation loss $\mathcal{L}_t$ and the segment-wise preference loss $\mathcal{L}_s$ as:
\begin{equation}\label{eq:seg-opd-objective}
\theta^*
=
\operatorname*{arg\,min}_{\theta}
\mathcal{L}_t(\theta)
+
\lambda\mathcal{L}_s(\theta),
\end{equation}
where $\lambda>0$ is the hyperparameter to control the contribution of the segment-wise preference loss.
Appendices~\ref{app:training} and \ref{app:complexity} detail \method's training algorithm and complexity analysis, respectively.

\section{Experiment}


In this section, we conduct extensive experiments to answer the following research questions:
\begin{itemize}[leftmargin=*,topsep=0pt,parsep=0pt]
    \item \textbf{RQ1:} How does \method perform against state-of-the-art baselines for on-policy distillation? 
    \item \textbf{RQ2:} How does \method affect the student's reasoning behavior?  
    \item \textbf{RQ3:} How effective are the proposed components of \method?
\end{itemize}

\subsection{Experiment Setup}
\noindent\textbf{Models}
We use DeepSeek-1.5B~\citep{guo2025deepseekr1}, Qwen3-4B, Qwen3-8B, and Qwen3-14B~\citep{yang2025qwen3} as students.
Each student is paired with an RL-trained teacher from the same model family and parameter scale.
Appendix~\ref{app:models} provides more details.

\noindent\textbf{Datasets}
We evaluate \method on mathematical reasoning and competitive programming.
For mathematical reasoning, we train on DeepMath-103K~\citep{he2025deepmath} and evaluate on AIME24, AIME25, AIME26~\citep{cai2026learning}, MATH500~\citep{lightman2024lets}, and MINERVA~\citep{lewkowycz2022solving}.
For competitive programming, we train on Eurus-RL-Code, evaluate on Codeforces, TACO~\citep{cui2025process}, and HumanEval+~\citep{liu2023evalplus}.
Appendix~\ref{app:datasets} provides more details.

\noindent\textbf{Baselines}
We compare \method against OPD~\citep {agarwal2024policy}, G-OPD with reward extrapolation~\citep{yang2026learning}, E-OPD, which adds a forward-KL constraint at the teacher's uncertain positions~\citep{jin2026entropy}, and OmniOPD~\citep{zhou2026omniopd}, which supervises student chunks through similarity with teacher continuations.
Appendix~\ref{app:baselines} provides more details.

\begin{figure}[t]
    \centering
    \includegraphics[width=\textwidth]{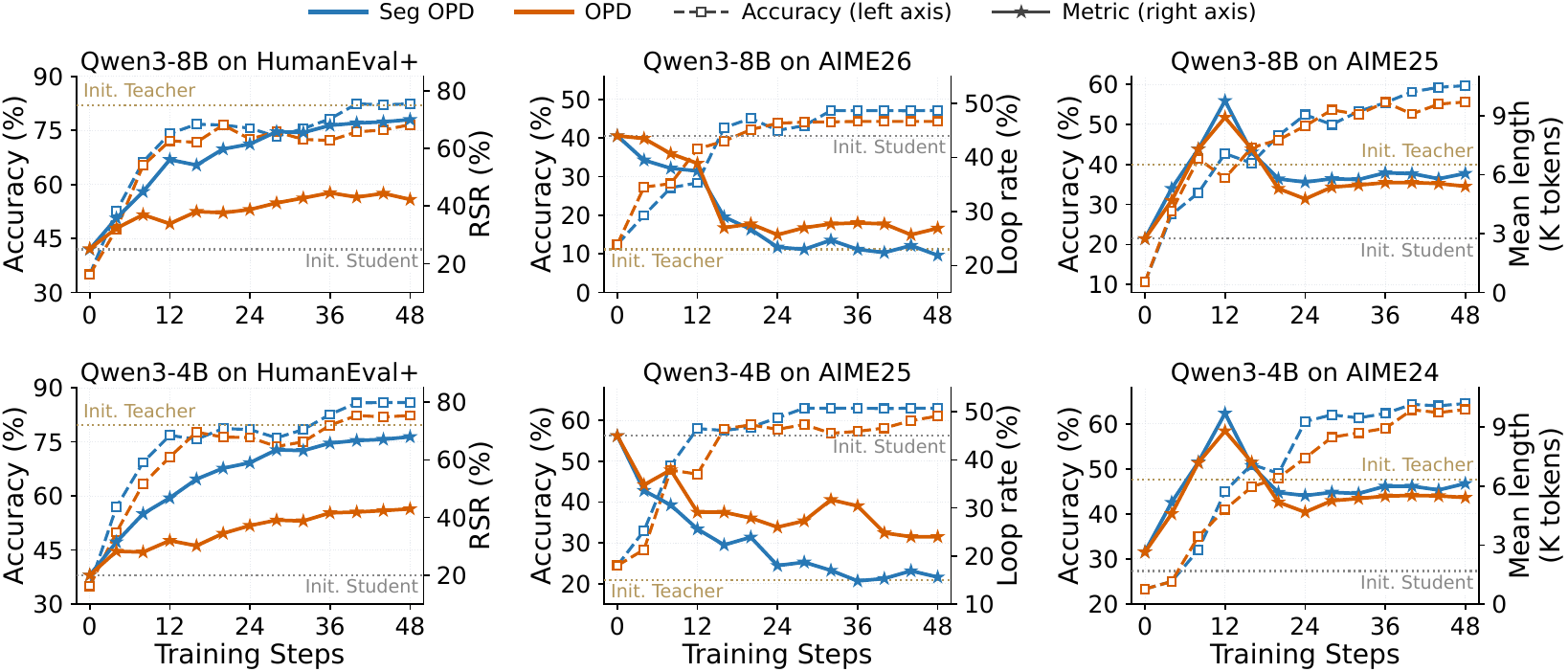}
    \caption{\textbf{Reasoning accuracy} (dashed lines, left axis) and \textbf{behavioral metric statistics} (solid lines, right axis). 
    \textcolor[HTML]{888888}{Gray} and \textcolor[HTML]{B2965B}{gold} horizontal lines indicate the initial behavioral metric values for the student and teacher models, respectively.
    }
    \label{fig:reasoning-behavior}
    \vspace{-1em}
\end{figure}

\subsection{Segment-wise OPD Performance}\label{sec:rq1}
To answer \textbf{RQ1}, we compare \method with competitive baselines on mathematical reasoning and competitive programming.
Results in Tab~\ref{tab:2} demonstrate that \method achieves an average relative accuracy gain of 5.22\% over baselines across all models and datasets.
Specifically, we observe that:
\begin{itemize}[leftmargin=*,topsep=0pt,parsep=0pt]
    \item Compared with standard OPD, \method improves reasoning accuracy in most settings, with gains of up to 8.24\% for DeepSeek-1.5B on AIME24.
    This improvement demonstrates the effectiveness of \method's segment-wise preference learning.

    \item \method outperforms E-OPD and G-OPD in most settings.
    Using student uncertainty, \method selects segments around reasoning pivots. It obtains teacher redrafts that offer alternative reasoning paths, giving students examples of how to redirect reasoning at potential turning points, rather than requiring students to infer such changes from token-wise distillation signals alone.

    
    \item \method also outperforms the closely related baseline OmniOPD in most settings, with gains of up to 8.83\% for DeepSeek-1.5B on AIME24.
    OmniOPD weights student chunks by their similarity to teacher continuations, whereas \method directly learns to prefer teacher redrafts over their paired student segments.

\end{itemize}

\subsection{Analysis of Student Reasoning Behavior}\label{sec:reasoning-behavior}

To answer \textbf{RQ2}, we compare \method with OPD, using Qwen3-8B and Qwen3-4B as students, and examine their reasoning behavior during training.
Alongside reasoning accuracy, we report the student's revision success rate (\textbf{RSR}) with the same definition as in Sec.~\ref{sec:casestudy}, the \textbf{loop rate}, defined as the proportion of steps that repeat earlier reasoning without progress, and \textbf{mean length}, the average number of tokens per rollout.
Appendix~\ref{app:behavior-analysis} provides further analysis on student reasoning behavior.
We make the following observations from Fig.~\ref{fig:reasoning-behavior}:
\begin{itemize}[leftmargin=*,topsep=0pt,parsep=0pt]
    \item Higher RSR with \method suggests that learning from teacher redrafts helps the student revise its own reasoning more successfully.
    For Qwen3-8B on HumanEval+, RSR peaks at 70.0\%, compared with 44.6\% for OPD and 75.0\% for the initial teacher.
    Peak reasoning accuracy reaches 82.4\% with \method versus 76.6\% with OPD.

    \item Lower loop rates accompany higher accuracy in training.
    For Qwen3-4B on AIME25, \method achieves a 15.6\% loop rate and 63.0\% accuracy, versus 24.0\% and 61.1\% with OPD, suggesting that revision learning helps students move past stalled reasoning toward a solution.

    \item For Qwen3-8B on AIME25, \method reaches 59.7\% accuracy, compared with 55.6\% for OPD, with mean response lengths of 6.1K and 5.4K tokens, respectively.
    Across both length comparisons, \method achieves higher final accuracy with comparable mean response lengths.

\end{itemize}

\subsection{Ablation Studies and Hyperparameter Sensitivity Analysis}\label{sec:rq3}

\begin{table}[t]
    \centering
    \small
    \setlength{\tabcolsep}{3pt}
    \renewcommand{\arraystretch}{1.05}
    \vspace{-0.3em}
    \caption{\textbf{Ablation studies} of \method components with Qwen3-4B as the student.}
    \vspace{-0.5em}
    \label{tab:ablation}
    \resizebox{\linewidth}{!}{
    \begin{tabular}{@{}lcccccccc@{}}
        \toprule
        \textbf{Variant} & \textbf{Codeforces} & \textbf{TACO} & \textbf{HumanEval+}& \textbf{AIME24} & \textbf{AIME25} & \textbf{AIME26} & \textbf{MATH500} & \textbf{MINERVA} \\
        \midrule
        \rowcolor{blue!8}
        \method & \textbf{45.89} & \textbf{43.65} & \textbf{85.89} & \textbf{64.50} & \textbf{63.00} & \textbf{52.53} & \textbf{89.10} & \textbf{48.85} \\
        \midrule
        \method\textbackslash{S} &  \secondbest{42.60} &  \secondbest{38.07} &  \secondbest{72.53} & 54.16 & \secondbest{59.18} & 51.54 & \secondbest{75.49} & 45.23 \\
        \method\textbackslash{T} & 40.27 & 33.25 & 71.89 & \secondbest{56.90} & 57.25 & \secondbest{52.39} & 70.05 & \secondbest{46.27} \\
        \method\textbackslash{N} & 37.58 & 31.75 & 70.69 & 42.96 & 56.89 & 49.20 & 73.29 & 40.67 \\
        \bottomrule
    \end{tabular}
    \vspace{-1.0em}
    }
\end{table}

\noindent\textbf{Ablation Studies.}
To answer \textbf{RQ3}, we evaluate three variants of \method:
\textbf{(i)} We replace pivot anchors with random anchors to obtain \method\textbackslash{S}.
\textbf{(ii)} We prompt the teacher in \method\textbackslash{T} to generate from the prefix $q$ without providing $s$ or explicit revision instruction.
\textbf{(iii)} We replace the segment-wise preference loss $\mathcal{L}_s$ in Eq.~(\ref{eq:segment-loss}) with supervised fine-tuning (SFT) loss on $r$ to obtain \method\textbackslash{N}.
From the results in Tab.~\ref{tab:ablation}, we observe that:

\begin{itemize}[leftmargin=*,topsep=0pt,parsep=0pt]
    \item Reasoning pivots provide more useful redrafting targets than position-matched random anchors, as shown by \method gains over \method\textbackslash{S}.

    \item The comparison with \method\textbackslash{T} shows that teacher redrafts tailored to the student's reasoning outperform teacher continuations from the prefix alone.

    \item The advantage over \method\textbackslash{N} supports learning to prefer teacher redrafts over their paired student segments, rather than simply imitating the redrafts.
    Appendix~\ref{app:training-dynamics} provides further analysis on the training dynamics and performance of \method and this variant.
\end{itemize}

\begin{figure}[t]
    \centering
    \includegraphics[width=\textwidth]{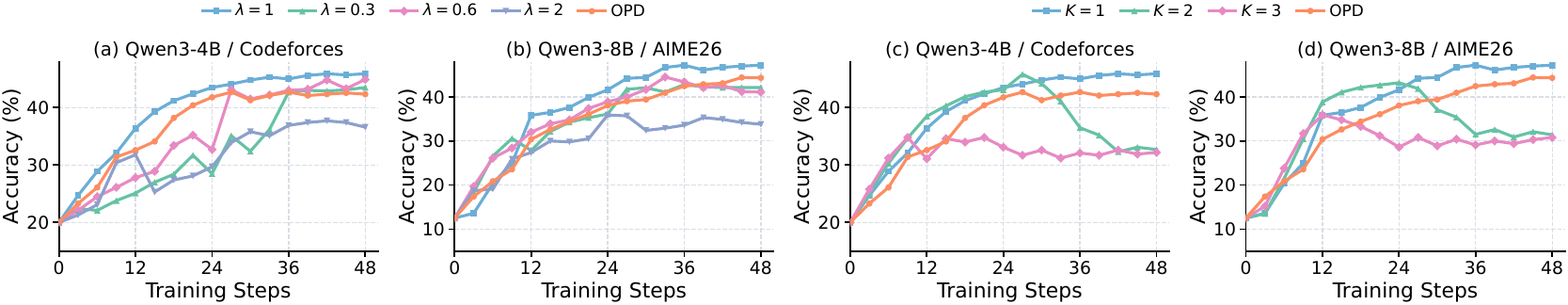}
    \caption{\textbf{Hyperparameter sensitivity analysis of \method.} 
    }
    \label{fig:hyperparameter-sensitivity}
    \vspace{-1em}
\end{figure}

\noindent\textbf{Hyperparameter Analysis.}
We vary the segment-wise loss weight $\lambda$ in Eq.~(\ref{eq:seg-opd-objective}) and the number $K$ of student segments selected per rollout in Eq.~\ref{eq:segment-uncertainty}.
Fig.~\ref{fig:hyperparameter-sensitivity} shows sweeps over $\lambda\in\{0.3,0.6,1,2\}$ and $K\in\{1,2,3\}$, with the default $\lambda$ used in Tab.~\ref{tab:2} normalized to $1$.
From the figure, we observe that:

\begin{itemize}[leftmargin=*,topsep=0pt,parsep=0pt]
    \item Smaller weights preserve accuracy, whereas $\lambda=2$ degrades both settings, ending at 33.8\% for Qwen3-8B on AIME26 versus 47.2\% at $\lambda=1$.
    This favors balancing revision supervision with token-wise distillation rather than simply increasing the preference weight.
    
    \item Larger $K$ weakens sustained performance, with accuracy at $K=2$ falling from a 45.8\% peak to 32.7\% by the end of training.
    Expanding the revision budget can upset the balance between segment-wise preference learning and token-wise distillation, making training less stable.
\end{itemize}

\section{Conclusion}
In this work, we address the limitations of standard OPD, where teacher feedback on degenerate student prefixes may reinforce problematic reasoning patterns without explicitly providing coherent alternative reasoning steps.
Our preliminary analysis shows that replacing student segments with teacher redrafts can improve subsequent reasoning accuracy.
Building on these findings, we develop an empirically grounded method for learning reasoning revision in OPD, with a learning objective that encourages students to prefer teacher redrafts over their paired segments.
Extensive experiments show that our \method consistently outperforms the competitive baselines in average reasoning accuracy across diverse models and tasks, while achieving higher revision success rates and fewer reasoning loops than baselines.
Ultimately, \method bridges the gap between teacher guidance on student segments and students' capacity for reasoning revision, enabling them to redirect subsequent reasoning when their current reasoning risks becoming problematic.


\clearpage

\newpage
\subsection*{AI use statement}


In this work, we used generative AI tools to illustrate the comic hiking metaphor in Fig.~\ref{fig:illustration} and to help identify and assess reasoning steps from the model's output.
Beyond the uses disclosed above, we did not use generative AI tools for any other tasks that required or recommended disclosure while completing this work.
We have reviewed all AI-assisted work by double-checking the output and revising it as needed. 
We take responsibility for the final content of this work, including text, claims, or artifacts produced with the aid of generative AI.


\subsection*{Ethics statement}


In this work, we explore reasoning revision learning to improve OPD methods.
We believe that our work conforms to the ICLR Code of Ethics. 
The proposed method, \method, helps improve understanding of the effectiveness of learning reasoning revision and makes LLM post-training improvement more accessible.
Despite the potential benefits, we acknowledge that more effective post-training techniques may also be misused to enhance or adapt LLMs for malicious purposes. 
We encourage responsible use of our method, together with the adoption of appropriate safeguards and ethical guidelines to mitigate potential risks.


\subsection*{Reproducibility statement}



In this work, we propose a novel approach to bridge the gap between teacher guidance on student rollouts and students' capacity for reasoning revision, enabling students to redirect subsequent reasoning when their current reasoning becomes problematic.
To ensure reproducibility, we provide comprehensive details throughout the paper and supplementary materials. 
The methodology of our proposed \method framework is described in detail in Section~\ref{sec:5}. 
Appendix~\ref{app:implementation} also provides a complete description of the implementation, including the datasets used, models trained, adaptations of baseline methods, all specific training parameters, and prompts we used in this work. 
Our code is also available in an anonymous repository. The link is provided at the end of the Abstract.


\bibliography{iclr2027_conference}
\bibliographystyle{iclr2027_conference}

\clearpage
\appendix
\raggedbottom \setlist[itemize]{itemsep=3pt,parsep=0pt,topsep=5pt,partopsep=0pt} \section{Implementation Details}\label{app:implementation}

\subsection{Details of RSR Evaluation}\label{app:rsr-evaluation}
In the main text, we propose a metric, revision success rate (RSR), in two places: \textbf{(i)}. We first introduce this metric in Sec.~\ref{sec:casestudy} to compare student revision behavior under random and pivot anchors, and in \textbf{(ii)} Sec.~\ref{sec:reasoning-behavior} to compare student revision behavior during training with \method and OPD.

In this subsection, we present details of RSR evaluation.
Manual identification of reasoning steps can be time-consuming and subjective.
To avoid this and apply a common evaluation procedure, we use GPT-5.6 Sol to identify reasoning steps in complete student rollouts, distinguish revision attempts from other steps, and assess whether each attempt corrects an earlier error with a substantively different alternative.
This process serves solely in RSR evaluation and differs from \method's boundary selection, since a reasoning step is not necessarily a multi-token span ended with a period, and vice versa.
Here we provide a prompt template for identifying reasoning steps and assessing revision success.

\begin{tcolorbox}[title={RSR evaluation prompt},colback=black!2,colframe=black!50,colbacktitle=black!50,coltitle=white,fonttitle=\small\bfseries,fontupper=\small\ttfamily,boxrule=0.4pt,arc=1mm,left=2mm,right=2mm,top=2mm,bottom=2mm,before upper={\raggedright\setlength{\parindent}{0pt}\setlength{\parskip}{0.5\baselineskip}}]
Given the problem and the student's complete response, identify all reasoning steps in their original order.

Prefer complete sentences where possible, but this is not required.

Preserve the original wording.

[Revision assessment]\\
For each step, determine whether it attempts to correct a suspected error in earlier reasoning.
For such an attempt, determine whether it corrects an error in an earlier step by providing a substantively different alternative.

Assess it in the context of the preceding reasoning.

Changes in wording or length alone do not count as successful revisions.

[Output]\\


Return every extracted step with a yes/no judgment on whether it is a revision attempt. For each revision attempt, also provide a yes/no judgment on whether it succeeds, with a brief but confirmative explanation.

[Problem]\\
\{problem\}

[Student response]\\
\{student\_response\}
\end{tcolorbox}

Simply put, a revision attempt is a reasoning step that seeks to correct a suspected error in earlier reasoning, whether it succeeds or not. 
We compute RSR as the number of successful revision attempts divided by the total number of revision attempts, including both successful and unsuccessful attempts. 
Steps that do not attempt a revision are excluded from the denominator.

Additionally, we highlight that RSR evaluation and training segment construction serve different purposes by:
\begin{itemize}[leftmargin=*]
    \item \textbf{RSR evaluation} identifies steps by their reasoning content to assess revisions in complete student rollouts.
    \item \textbf{Training segment construction} uses construction boundaries around pivot anchors to construct local segments for teacher redrafting, without using the RSR evaluator to determine these boundaries.
\end{itemize}
The preference for complete sentences in evaluation does not require its step boundaries to coincide with those of the training segments.

\subsection{Details of Revision Unit Construction}\label{app:segment}
To provide coherent revision targets for teacher redrafting, we propose to extract complete sentences around reasoning pivots.
Following the use of periods for step segmentation~\citep{golovneva2023roscoe,wang-etal-2025-offline}, we use periods to identify sentence boundaries.
For each pivot anchor $t_i\in T$ in a failed student rollout $y$, we search backward and forward for the surrounding periods and extract the enclosing sentence as $s_i=y_{s_i^{b}:s_i^{e}}$, where $s_i^{b}$ and $s_i^{e}$ index its first and last tokens.
We score the candidates using Eq.~(\ref{eq:segment-uncertainty}) and greedily select non-overlapping segments in descending score order, up to the per-rollout budget $K$.

For each selected student segment $s$, we seek an alternative reasoning step under the same prefix $q$.
We provide $q$ and $s$ to the fixed teacher, prompt it to replace $s$ with its own reasoning, and let it continue until it solves the problem.
From the resulting full draft $\tilde r$, we retain the leading sentence ending at the first period as the teacher redraft $r$ only if $\tilde r$ passes the task-specific correctness check.
Applying the same sentence boundary rule to both branches yields the revision unit $p=(q,s,r)$ for segment-wise preference learning.
We send each prompt below as a single user message with thinking disabled.

Specifically, the prompt fields \texttt{\{problem\}} and \texttt{\{q\}} contain the problem and the preceding student response, respectively, which together form the conditioning prefix $q$; \texttt{\{s\_initial\}} contains student segment $s$.
We generate full teacher drafts with temperature $0.2$ and top-$p$ $0.9$, allowing up to $8{,}192$ new tokens within a $32{,}768$-token context window.
We discard sentence candidates without a complete right boundary, as well as empty teacher drafts and drafts truncated at the generation limit.

\begin{tcolorbox}[title={Teacher redrafting prompt for mathematical reasoning},colback=black!2,colframe=black!50,colbacktitle=black!50,coltitle=white,fonttitle=\small\bfseries,fontupper=\small\ttfamily,boxrule=0.4pt,arc=1mm,left=2mm,right=2mm,top=2mm,bottom=2mm,before upper={\raggedright\setlength{\parindent}{0pt}\setlength{\parskip}{0.5\baselineskip}}]
Continue the mathematical solution from the exact point shown below.

[Problem]\\
\{problem\}

[Solution prefix]\\
\{q\}

[Student next step to replace; do not quote it]\\
\{s\_initial\}

[Task]\\
Write a mathematically correct continuation beginning at exactly the same point as the student step. Start directly with solution text: do not discuss the prompt, the student, or your instructions. Continue naturally until the problem is fully solved, including the final answer.
\end{tcolorbox}

\begin{tcolorbox}[title={Teacher redrafting prompt for programming},colback=black!2,colframe=black!50,colbacktitle=black!50,coltitle=white,fonttitle=\small\bfseries,fontupper=\small\ttfamily,boxrule=0.4pt,arc=1mm,left=2mm,right=2mm,top=2mm,bottom=2mm,before upper={\raggedright\setlength{\parindent}{0pt}\setlength{\parskip}{0.5\baselineskip}}]
Continue the Python solution from the exact point shown below.

[Problem]\\
\{problem\}

[Solution prefix]\\
\{q\}

[Student next step to replace; do not quote it]\\
\{s\_initial\}

[Task]\\
Write a corrected continuation beginning at exactly the same point and programming subgoal as the student step. Start directly with solution text: do not discuss the prompt, the student, or your instructions. Continue naturally until the problem is fully solved, ending with one complete executable Python code block.
\end{tcolorbox}

To assess the quality of revision units, we conduct an audit with the following settings:
\begin{itemize}[leftmargin=*,topsep=2pt,itemsep=2pt]
\item We randomly sample 1,024 revision units generated from a Qwen3-4B and Qwen3-8B student-teacher pair on mathematical reasoning and programming.
\item We use the GPT-5.6 Sol model to evaluate the prefix $q$, the retained segments $s$, and the redraft $r$, presented anonymously in random order without revealing the sources. 
\item We assess whether $s$ and $r$ can be evaluated as local alternatives under $q$, then compare their correctness and progress toward a solution. Different valid reasoning strategies are allowed.
\item We also label pairs as ties when evidence is insufficient to indicate a clear preference.
\end{itemize}

\begin{table}[t!]
\centering
\caption{Revision unit quality assessment.}
\small
\setlength{\tabcolsep}{4pt}
\label{tab:pair-quality-audit}
\begin{tabular}{llcc}
\toprule
\textbf{Student Model} & \textbf{Audit results} & \textbf{Math (\%)} & \textbf{Code (\%)} \\
\midrule
\multirow{3}{*}{Qwen3-4B} & Redraft preferred & 86.9 & 89.7 \\
 & Segment preferred & 2.6 & 6.4 \\
 & Tie & 10.5 & 3.9 \\
\midrule
\multirow{3}{*}{Qwen3-8B} & Redraft preferred & 87.3 & 88.4 \\
 & Segment preferred & 3.2 & 5.9 \\
 & Tie & 9.5 & 5.7 \\
\bottomrule
\end{tabular}
\end{table}

We record the results in Tab.~\ref{tab:pair-quality-audit}. From the table, we have the following key observations:
\begin{itemize}[leftmargin=*]
\item The evaluator prefers teacher redrafts in 86.9\%--89.7\% of audited units, compared with 2.6\%--6.4\% for student segments. This pattern holds for both Qwen3-4B and Qwen3-8B in mathematics and programming, supporting the training preference $r \succ s$ under the audit's criteria of correctness and progress toward a solution.
\item We assess the preference from the retained segments under $q$, without access to the full teacher draft. These judgments support using short redrafts as local preference targets based on the retained text. Ties remain in the denominator, so the redraft preference rates cover all audited units, including those without a clear winner.
\end{itemize}

\subsection{Details of Models}\label{app:models}

\begin{table}[t!]
\centering
\caption{Student--teacher pairs. The algorithm column refers to teacher RL training algorithms.}
\small
\setlength{\tabcolsep}{4pt}
\label{tab:models}
\begin{tabular}{lcc}
\toprule
\rowcolor{gray!12}
\multicolumn{3}{c}{\textit{Mathematical Reasoning}} \\
\midrule
\textbf{Student Model} & \textbf{Teacher Model} & \textbf{Algorithm} \\
\midrule
DeepSeek-1.5B & JustRL-DeepSeek-1.5B~\citep{he2025justrlscaling15bllm} & GRPO \\
Qwen3-4B & Qwen3-4B-Non-Thinking-RL-Math~\citep{yang2026learning} & GRPO \\
Qwen3-8B & Qwen3-8B-DAPO~\citep{cai2025predictability} & DAPO \\
Qwen3-14B &  Qwen3-14B-Math-RL~\citep{cai2026learning} & DAPO \\
\midrule
\rowcolor{gray!12}
\multicolumn{3}{c}{\textit{Competitive Programming}} \\
\midrule
\textbf{Student Model} & \textbf{Teacher Model} & \textbf{Algorithm} \\
\midrule
DeepSeek-1.5B & JustRL-DeepSeek-1.5B~\citep{he2025justrlscaling15bllm} & GRPO \\
Qwen3-4B & Qwen3-4B-Non-Thinking-RL-Code~\citep{yang2026learning} & GRPO \\
Qwen3-8B & Qwen3-8B-Code-RL~\citep{cai2025predictability} & DAPO \\
Qwen3-14B & Qwen3-14B-Code-RL~\citep{cai2026learning} & GRPO \\
\bottomrule
\end{tabular}
\end{table}

To highlight the scalability of our \method and the generalizability of our findings, we conduct extensive experiments across models with diverse parameter scales, ranging from 1.5B to 14B parameters.
Our student models include publicly available pre-trained and fine-tuned checkpoints, as well as models that we locally trained using the \texttt{verl} framework.

For all student models reported in the main text, the corresponding teacher model is consistently the RL-trained version of its base model.
This setting aligns with~\citep{cai2026learning}, which emphasizes using the same model architecture for both teacher and student models to ensure a fair comparison and stable training.
Tab.~\ref{tab:models} lists the student--teacher pairs for mathematical reasoning and competitive programming, together with the teachers' RL training algorithms and sources.

\subsection{Details of Datasets}\label{app:datasets}

\begin{table}[t!]
\centering
\caption{The statistics of datasets in our work.}
\small
\setlength{\tabcolsep}{6pt}
\renewcommand{\arraystretch}{1.05}
\label{tab:datasets}
\begin{tabular}{ll>{\raggedright\arraybackslash}p{0.40\textwidth}}
\toprule
\textbf{Evaluation Dataset} & \textbf{Size} & \textbf{Description} \\
\midrule
\rowcolor{gray!12}
\textit{Training on DeepMath-103K}~\citep{he2025deepmath} & & \\
\midrule
AIME24 &  & \multirow{3}{=}{American Invitational Mathematics Examination from year 2024 to 2026.} \\
AIME25 & 30 & \\
AIME26~\citep{cai2026learning} &  & \\
\midrule
MATH500~\citep{lightman2024lets} & 500 & MATH subset with varied difficulty \\
MINERVA~\citep{lewkowycz2022solving} & 272 & Undergraduate math and science \\
\midrule
\rowcolor{gray!12}
\textit{Training on Eurus-RL-Code}~\citep{cui2025process} & & \\
\midrule
Codeforces~\citep{cui2025process} & 123 & Competitive programming \\
TACO~\citep{cui2025process} & 382 & Diverse algorithmic problems \\
HumanEval+~\citep{liu2023evalplus} & 164 & Python functions with expanded tests \\
\bottomrule
\end{tabular}
\end{table}

In this work, we evaluate mathematical reasoning and competitive programming.
Following~\citep{cai2026learning}, we fully leverage domain-specialized capabilities by training the model on DeepMath-103K~\citep{he2025deepmath} for mathematical reasoning and Eurus-RL-Code~\citep{cui2025process} for competitive programming.
After filtering, the training datasets contain 57,046 DeepMath-103K problems and 25,276 Eurus-RL-Code problems, covering a wide range of topics and difficulty levels.
We evaluate our \method and competitors on extensive datasets that vary in source, scale, and difficulty to highlight the transferability of different methods.
Table~\ref{tab:datasets} summarizes the datasets used in our experiments.

\subsection{Details of Compared Methods}\label{app:baselines}
In the main text, we compare \method with representative and state-of-the-art OPD methods.
For a fair comparison, we reproduce all baselines in the same \texttt{verl} framework, using the student--teacher pairs, training data, and evaluation protocol described above.
We tune each method's hyperparameters based on student performance on a validation split held out from the training dataset and select its best validation configuration.
This approach ensures that we evaluate each method under its best-performing conditions. 
For instance, G-OPD with Qwen3-4B achieves 61.2\% on AIME25 under our experimental setting, compared with 56.1\% best performance reported in the original paper~\citep{yang2026learning}.
Here we present the methods and their implementations as follows:
\begin{itemize}[leftmargin=*]
    \item \textbf{OPD}~\citep{agarwal2024policy,yang2026learning}: This foundational OPD method trains the student on its own generated trajectories using token-level feedback from the teacher.
    We reproduce the sampled-token reverse-KL objective, with teacher log-probabilities evaluated at every response token.

    \item \textbf{E-OPD}~\citep{jin2026entropy}: E-OPD augments standard OPD with forward-KL supervision at positions where the teacher has high entropy, encouraging the student to retain multiple plausible continuations.
    We adapt the authors' released implementation to our training framework, preserving the teacher-entropy gate and the forward-KL term over the teacher's top-$k$ tokens.

    \item \textbf{G-OPD}~\citep{yang2026learning}: G-OPD scales the implicit reward defined by the log-probability difference between the teacher and a reference model.
    We reproduce this method with a reward extrapolation coefficient of 1.25 and use a frozen copy of the initial student as the reference model.

    \item \textbf{OmniOPD}~\citep{zhou2026omniopd}: OmniOPD weights entropy-selected student chunks by their agreement with teacher continuations, using only text outputs from the teacher.
    We reproduce its Bayesian chunk weighting and KL regularization from the paper, using ROUGE-1 F1~\citep{lin2004rouge} to measure agreement and the frozen initial student to regularize unaudited tokens.
\end{itemize}

\subsection{Other Implementation Details}\label{app:other}

We implement \method and all baselines based on \texttt{PyTorch}~2.10.0, \texttt{verl}~0.8.0~\citep{sheng2024hybridflow}, \texttt{vLLM}~0.18.0~\citep{kwon2023efficient}, and \texttt{Transformers}~5.3.0~\citep{wolf-etal-2020-transformers}.
All the experiments are conducted on 8 $\times$ NVIDIA H100 80 GB GPUs.


\begin{itemize}[leftmargin=*,topsep=2pt,itemsep=2pt,parsep=0pt]
    \item \textbf{Optimizer:} AdamW with $(\beta_1,\beta_2)=(0.9,0.999)$ and no weight decay.
    \item \textbf{Learning rate:} $5\times10^{-6}$ with cosine decay and without warmup.
    \item \textbf{Rollout batch size:} $256$ for Qwen3-14B and $512$ for the other students.
    \item \textbf{Student rollout sampling:} One rollout per problem, with temperature $1.0$ and top-$p$ $1.0$, without top-$k$ filtering.
    \item \textbf{Maximum prompt length:} $2,048$ tokens.
    \item \textbf{Student response limit:} $22,000$ tokens for mathematical reasoning and $30,000$ tokens for competitive programming.
    \item \textbf{Seeds:} We use $42$ by default. For repeated runs, we use additional randomly selected seeds.
    \item \textbf{Segment budget:} $K=1$ unless otherwise specified.
    \item \textbf{Preference margin scale:} $\beta=0.01$.
    \item \textbf{Segment-wise loss weight:} For mathematical reasoning, $\lambda$ is set as $\{0.75,0.35,0.33,0.5\}$ for DeepSeek-1.5B, Qwen3-4B, Qwen3-8B, and Qwen3-14B, respectively. For competitive programming, $\lambda$ is set as $\{0.67,0.9,0.67,0.75\}$ for DeepSeek-1.5B, Qwen3-4B, Qwen3-8B, and Qwen3-14B, respectively.
    \item \textbf{Training steps:} $48$.
    \item \textbf{Training time validation frequency:} Every $4$ steps unless otherwise specified.
\end{itemize}

At each checkpoint, we generate $4$ responses per problem with temperature $1.0$ and top-p $1.0$, averaging correctness first over responses and then over problems.
For all methods, we report mean Pass@1 from the final checkpoint after all training steps as reasoning accuracy, unless otherwise specified, such as peak values that refer to the highest observed values of the stated metrics over the displayed checkpoints.
We provide an example training command for \method in the next box.



\begin{tcolorbox}[title={\method training command},colback=black!2,colframe=black!50,colbacktitle=black!50,coltitle=white,fonttitle=\small\bfseries,fontupper=\scriptsize,boxrule=0.4pt,arc=1mm,left=2mm,right=2mm,top=1mm,bottom=1mm]
\begin{verbatim}
python3 -m verl.trainer.main_ppo \
  data.train_files="$TRAIN_FILE" \
  data.val_files="$VAL_FILES" \
  data.train_batch_size=512 \
  data.max_prompt_length=2048 \
  data.max_response_length=22000 \
  data.seed=42 \
  +data.apply_chat_template_kwargs.enable_thinking=False \
  actor_rollout_ref.model.path="$STUDENT_MODEL" \
  actor_rollout_ref.actor.optim.lr=5e-6 \
  actor_rollout_ref.actor.optim.lr_scheduler_type=cosine \
  actor_rollout_ref.actor.optim.lr_warmup_steps_ratio=0.0 \
  actor_rollout_ref.actor.ppo_mini_batch_size=128 \
  actor_rollout_ref.actor.ppo_epochs=1 \
  actor_rollout_ref.actor.policy_loss.seg_opd_enable=True \
  actor_rollout_ref.actor.policy_loss.seg_opd_token_loss_coef=1.0 \
  actor_rollout_ref.actor.policy_loss.seg_opd_segment_loss_coef=1.0 \
  actor_rollout_ref.actor.policy_loss.seg_opd_dpo_beta=0.01 \
  actor_rollout_ref.actor.policy_loss.seg_opd_dpo_neutral_gate_cap=False \
  actor_rollout_ref.rollout.n=1 \
  actor_rollout_ref.rollout.temperature=1.0 \
  actor_rollout_ref.rollout.top_p=1.0 \
  actor_rollout_ref.rollout.top_k=-1 \
  +trainer.seg_opd_teacher_model_path="$TEACHER_MODEL" \
  +trainer.seg_opd_max_segments_per_rollout=1 \
  +trainer.seg_opd_full_redraft_max_tokens=8192 \
  +trainer.seg_opd_redraft_temperature=0.2 \
  +trainer.seg_opd_redraft_top_p=0.9 \
  trainer.n_gpus_per_node=8 \
  trainer.nnodes=1 \
  trainer.test_freq=4 \
  trainer.total_training_steps=48
\end{verbatim}
\end{tcolorbox}

\section{Additional Experiments and Analysis}

\subsection{Additional Analysis on Boundary Construction Rules}
\label{app:segment-selection-analysis}

\begin{figure}[!htbp]
    \centering
    \includegraphics[width=\linewidth]{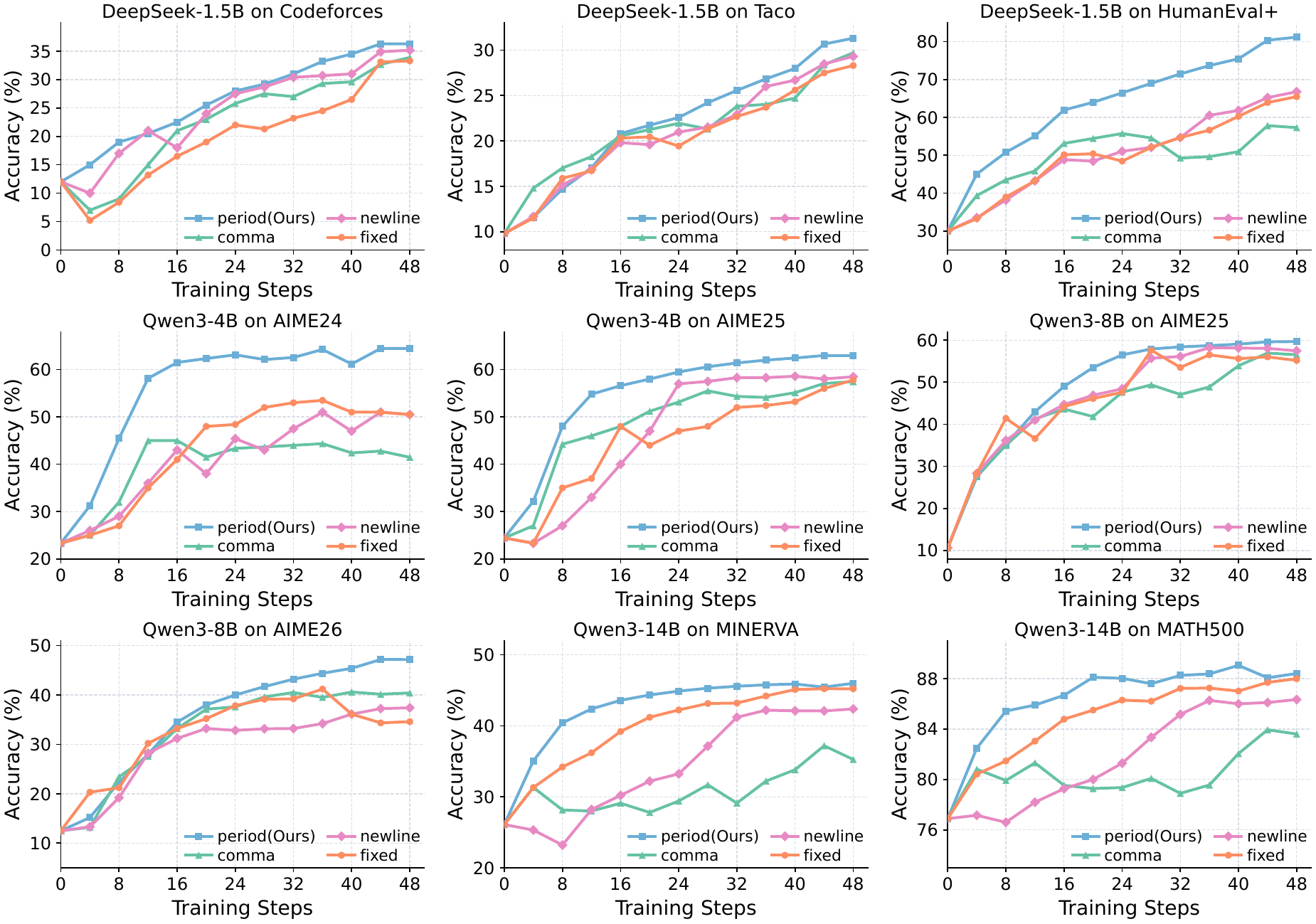}
    \caption{Reasoning accuracy of \method with four different construction rules. 
    }
    \label{fig:app-segment-boundaries}
\end{figure}

In the main text, we propose selecting boundaries from student rollouts to form segments.
With selected reasoning pivots, \method extends the boundaries from them in both directions until a full segment is formed.
Primarily, we aim to build segments and redrafts that are both integrated and representative of the underlying reasoning process of the models, requiring us to preserve the completeness of the segment while ensuring it is informative.
Punctuation and line breaks provide explicit structural cues in model outputs, allowing simple rules to guide both student segment construction and teacher redraft extraction.
To achieve this, we conduct extensive analysis to assess the effectiveness of various construction rules in the context of our \method.
Specifically, we examine four different construction rules. We extend the boundaries from the selected pivots until we encounter the first \textbf{(i)} periods, \textbf{(ii)} commas, \textbf{(iii)} newlines, and \textbf{(iv)} a fixed length of $2,048$ tokens.
We record the results for each construction rule across diverse models and datasets in Fig.~\ref{fig:app-segment-boundaries}.
From the figure, we have the following observations:

\begin{itemize}[leftmargin=*]
\item \textbf{Period} rule performs best overall, remaining at or near the top late in training.
Period boundaries preserve sentence-level context for teacher redrafting.
This helps retain a complete local inference and its explanation in the paired texts, making the reasoning contrast between student segments and teacher redrafts clearer for preference learning.

\item \textbf{Comma} rule generally trails the period, with a clear gap on HumanEval+.
Comma boundaries may fragment a reasoning step, separating an operation from its justification and obscuring the reasoning contrast between the student segment and teacher redraft.

\item \textbf{Newline} rule approaches period accuracy on Codeforces but remains weaker on HumanEval+.
Prior work uses double line breaks for step segmentation~\citep{zhang2025lessons}, but line breaks can also reflect layout rather than reasoning boundaries.

\item \textbf{Fixed} rule approaches the accuracy of that of the period rule on MINERVA but underperforms it on AIME26.
Fixed token cuts may truncate a step or include later reasoning, especially when the student and teacher express the same step at different lengths.
\end{itemize}

\subsection{Additional Analysis on Reasoning Behavior}\label{app:behavior-analysis}
\label{app:reasoning-behavior}

To extend the analysis in Sec.~\ref{sec:reasoning-behavior}, we examine how revision learning changes the student's use of reasoning transition markers and whether its RSR gains extend to additional model--dataset settings and a closely related baseline.

We report \textbf{word ratio}, the proportion of English words in natural-language reasoning that belong to our predefined set, including \texttt{wait}, \texttt{however}, \texttt{therefore}, \texttt{check}, \texttt{verify}, \texttt{correct}, \texttt{error}, \texttt{mistake}, and \texttt{incorrect}.
Code is excluded from this calculation.
Prior work uses related expressions as lexical indicators of reflection, backtracking, and self-verification~\citep{huang2026learning}.
Motivated by this practice, we use word ratio as a lexical proxy for potential reasoning transitions and additionally report \textbf{RSR}, as defined in Sec.~\ref{sec:casestudy}, to assess successful revisions.

\begin{figure}[!t!]
    \centering
    \includegraphics[width=\linewidth]{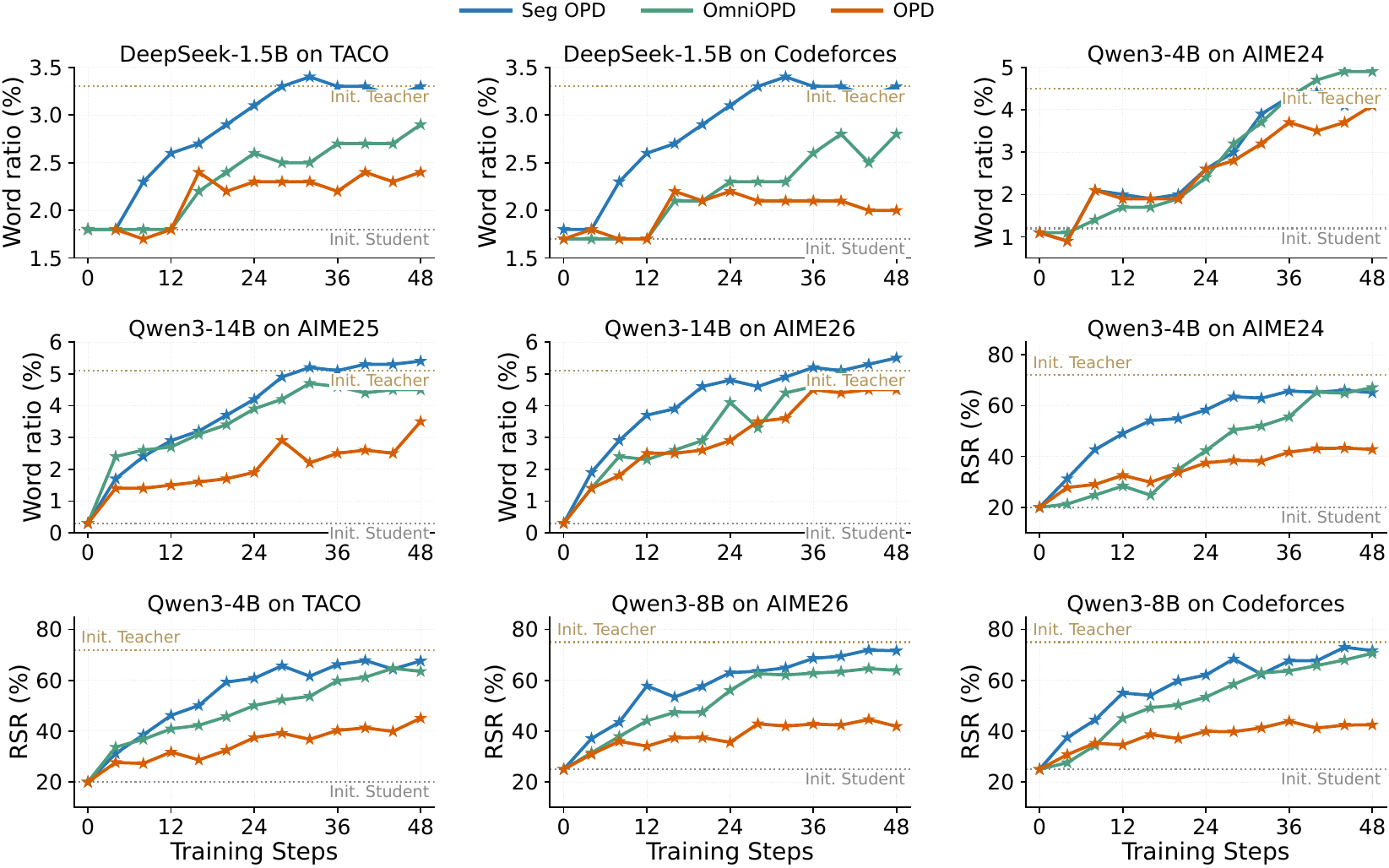}
    \caption{Additional analysis of reasoning behavior of \method, OmniOPD, and OPD. 
    The first five panels report word ratio, and the remaining four report revision success rate (RSR), ordered from left to right and top to bottom. 
    Horizontal lines indicate the initial behavioral metric values for the student and teacher models, respectively.
    }
    \label{fig:app-behavior-analysis}
\end{figure}

In addition to OPD, we include OmniOPD~\citep{zhou2026omniopd}, a closely related baseline that also applies teacher supervision to multi-token chunks of student-generated trajectories.
OmniOPD uses similarity with teacher continuations to weight learning from student chunks, whereas \method directly learns to prefer teacher redrafts over paired student segments.
We make the following observations from Fig.~\ref{fig:app-behavior-analysis}:
\begin{itemize}[leftmargin=*]
    \item \method increases the use of transition markers across the displayed settings, with higher word ratios than both baselines late in training for DeepSeek-1.5B and Qwen3-14B.
    For DeepSeek-1.5B, the ratio approaches the initial teacher's level on both TACO and Codeforces.
    This reflects increased use of the selected reasoning-related expressions.

    \item Late in training, \method attains higher RSR than OPD in all four additional settings, covering both mathematical and competitive programming.
    Successful revisions therefore account for a larger proportion of the student's revision attempts across these settings, extending the behavioral gains observed in the main text.

    \item Compared with OmniOPD, \method generally reaches higher RSR earlier in training, while the gap later narrows on Qwen3-4B/AIME24 and Qwen3-8B/Codeforces.
    On Qwen3-8B/AIME26, a clear gap remains late in training.
    These trajectories show earlier RSR gains with \method than with OmniOPD, even when their late-stage RSR becomes similar.
\end{itemize}

\subsection{Additional Analysis on Training Dynamics of \method}\label{app:training-dynamics}

\begin{figure}[t!]
    \centering
    \includegraphics[width=\linewidth]{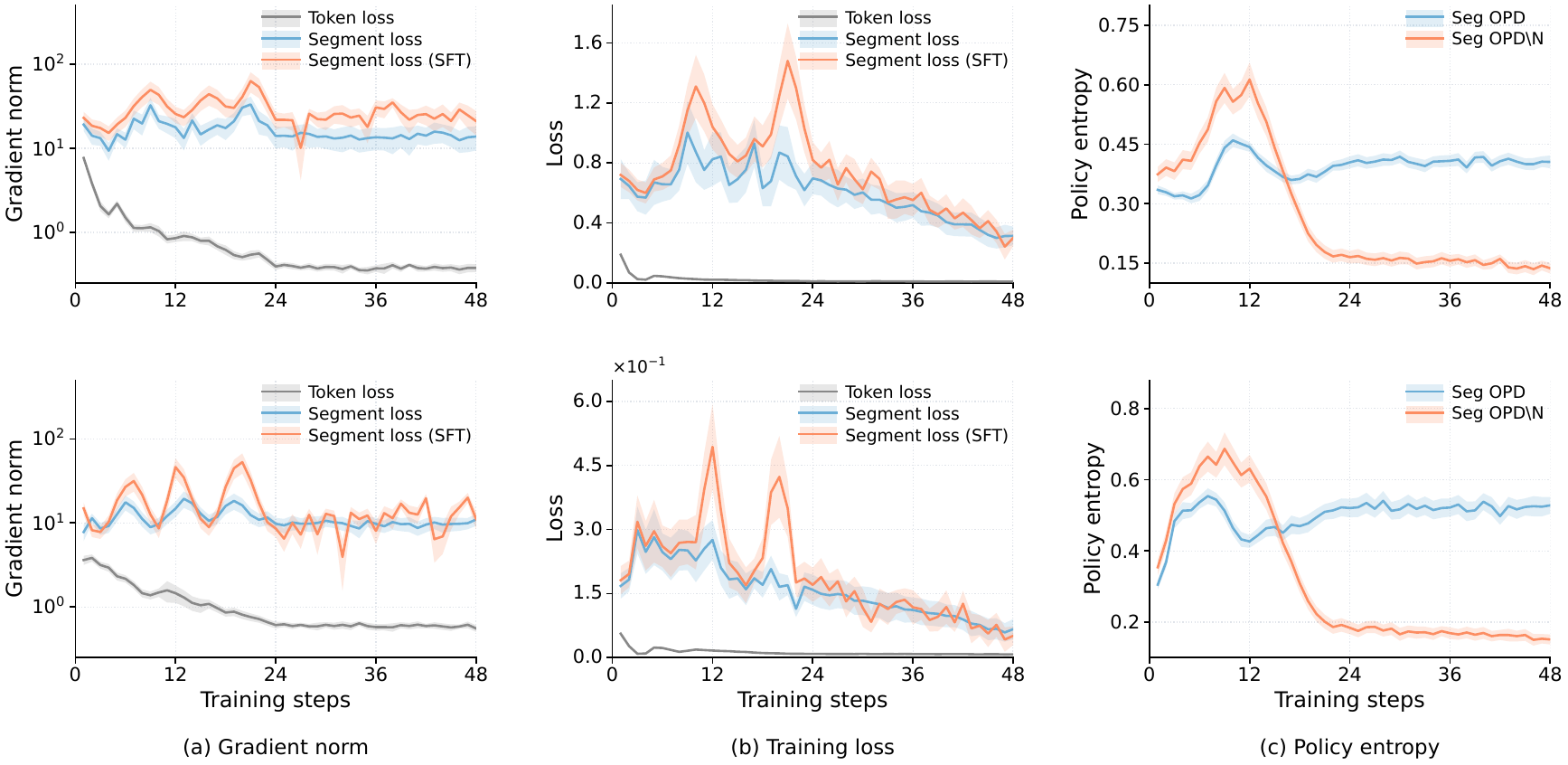}
    \caption{Training dynamics of training a Qwen3-4B on DeepMath-103K \textbf{(first row)} and Eurus-RL-Code \textbf{(second row)}. 
    \textbf{Left} column and \textbf{middle} columns compare the gradient norms and values of the token loss, segment loss, and segment loss (SFT) used in \method\textbackslash{N} variant, respectively.
    \textbf{Right} column compares the policy entropy of \method and its SFT variant.
    We record the mean and standard deviation across five runs with different random seeds.
    }
    \label{fig:app-training-dynamics}
\end{figure}

We investigate \method's training dynamics by tracking its \textbf{(i)} loss gradient norms, \textbf{(ii)} training losses, and \textbf{(iii)} policy entropy.
We compute gradient norms with respect to model parameters before clipping, and report component loss values before applying their mixture coefficients.
Policy entropy measures the student's average uncertainty over the next token when generating responses.

Additionally, to examine the benefits of segment-wise preference learning over redraft imitation, we compare \method with its SFT variant \method\textbackslash{N} in Sec.~\ref{sec:rq3}.
This variant replaces the segment-wise preference loss with SFT on teacher redrafts while retaining token-wise OPD.
Specifically, we rewrite the segment-wise term $\mathcal{L}_s$ in Eq.~(\ref{eq:seg-opd-objective}) with a cross-entropy loss on teacher redrafts:

\begin{equation}\label{eq:segment-sft}
\mathcal{L}_s(\theta)
= -\frac{1}{N}\sum_{i=1}^{N}\frac{1}{|r_i|}
\sum_{t=1}^{|r_i|}
\log \pi_\theta\!\left(r_{i,t}\mid q_i,r_{i,<t}\right),
\end{equation}
where $N$ is the number of revision units in the training batch and $|r_i|$ is the number of tokens in the retained teacher redraft $r_i$.
The student policy $\pi_\theta$, with parameters $\theta$, predicts token $r_{i,t}$ given the prefix $q_i$ and preceding redraft tokens $r_{i,<t}$.
We first average the token losses within each redraft and then average across revision units, giving each unit equal weight in the loss.
This ablation replaces segment-wise preference learning with direct teacher-redraft imitation while retaining token-wise OPD.
The overall objective remains $\mathcal{L}_t+\lambda\mathcal{L}_s$, where $\mathcal{L}_t$ is the token-wise OPD loss, and $\lambda$ controls the relative strength of redraft imitation.
The SFT term omits the preference scale $\beta$ and the comparison against the student segment; thus, its gradient scale need not match that of the preference loss.
The coefficient $\lambda$ provides an overall rescaling of the SFT contribution.


We build on direct preference optimization (DPO), which learns from the contrast between preferred and rejected responses~\citep{rafailov2023direct}.
In \method, this contrast teaches the student to favor a teacher redraft over the student segment it revises, rather than only increasing the redraft's likelihood.
We normalize each segment's log policy ratio by its token length and rescale the difference by the mean segment length, as defined in Eq.~(\ref{eq:revision-margin}).

Fig.~\ref{fig:app-training-dynamics} shows the training dynamics of Qwen3-4B on DeepMath-103K and Eurus-RL-Code.
We make the following observations:
\begin{itemize}[leftmargin=*]
    \item \textbf{(i) Gradient norms.}
    The token-loss gradient norm generally decreases over training, while the segment-loss gradient norm continues to exhibit local fluctuations.
    Compared with SFT on teacher redrafts, the preference loss produces smaller gradient norms at most training steps, with lower peaks in both tasks.
    The larger SFT norms occur under per-redraft token averaging, not unnormalized token summation.
    Equal loss weight across revision units does not imply matched gradient scales between SFT and preference learning.

    \item \textbf{(ii) Training loss.}
    The token loss decreases in both tasks, indicating closer matching to the teacher's token predictions on sampled student trajectories.
    After early fluctuations, both the segment-wise preference loss and the SFT loss generally decrease.
    The preference loss penalizes small or negative revision margins, whereas the SFT loss measures how well the student predicts teacher redrafts.
    By explicitly favoring teacher redrafts over their paired student segments, \method uses the contrast between two reasoning alternatives as a direct training signal.
    This teaches the student which local reasoning choice to prefer under the same prefix, complementing token-wise distillation with supervision tailored to its own reasoning.
    The accuracy gains over redraft SFT in Tab.~\ref{tab:ablation} further support the value of learning from this contrast.

    \item \textbf{(iii) Policy entropy.}
    In both tasks, the SFT variant's policy entropy rises initially and then declines sharply, whereas \method maintains higher entropy in later steps.
    Lower entropy corresponds to more concentrated next-token predictions on the sampled trajectories.
    The SFT variant also scores below \method on all eight benchmarks in Tab.~\ref{tab:ablation}, with accuracy gaps ranging from 3.33\% on AIME26 to 21.54\% on AIME24.
    Together with the accuracy differences in Tab.~\ref{tab:ablation}, these observations support segment-wise preference learning over direct redraft imitation in this OPD setting.
\end{itemize}

\subsection{Case Study of Reasoning Correction}\label{app:revision-case}

In this subsection, we present two case studies of reasoning correction in student reasoning trajectories. Specifically, we organize these cases with the following settings:

\begin{itemize}[leftmargin=*,topsep=2pt,itemsep=2pt]
\item We present qualitative comparisons between the initial and \method-trained student responses to the same mathematical reasoning and competitive programming problems, using saved evaluation rollouts.
\item We highlight the same specific reasoning error in each response pair and show how the trained student subsequently corrects it.
\item We report the final mathematical answers and the trained student's programming test results.
\end{itemize}

For the mathematical reasoning case, we select AIME26 Problem \#4, which reduces a counting problem to finding distinct factors greater than one.
The key distinction is between identifying an equal-factor decomposition and showing that no admissible distinct-factor decomposition exists.
We focus on how the trained response checks a proposed exclusion rule: a correct algebraic reduction does not by itself justify excluding an entire class of numbers.

For the competitive programming case, we select TACO Problem \#797C, where both responses share the stack representation.
The key decision is when to pop the stack: once a character is appended to the output, later moves cannot change its position.
We focus on how the response tests the proposed greedy rule against the given example and reconsiders which information from the unread input should guide that decision.

We present the detailed cases on the next pages.

\begingroup
\newcommand{\caseannotation}[1]{\textcolor{blue!50!black}{\textsf{\textbf{#1}}}}
\tcbset{colback=black!4,colframe=black!75,colbacktitle=black!75,coltitle=white,fonttitle=\small\bfseries,fontupper=\small,boxrule=0.5pt,arc=1mm,left=2.5mm,right=2.5mm,top=1.5mm,bottom=1.5mm,before skip=5pt,after skip=5pt,before upper={\setlength{\parindent}{0pt}\setlength{\parskip}{0.2\baselineskip}},casestep/.style={colback=white,colframe=blue!50!black,boxrule=0.6pt,arc=0.7mm,left=1.6mm,right=1.6mm,top=1mm,bottom=1mm,before skip=4pt,after skip=4pt}}

\noindent\begin{minipage}{\linewidth}
\textbf{A mathematical reasoning case}\par\smallskip
\begin{tcolorbox}[title={Question (AIME26, Problem \#4)}]
Find the number of integers less than or equal to 100 that are equal to $a+b+ab$ for some choice of distinct positive integers $a$ and $b$.
\end{tcolorbox}

\begin{tcolorbox}[title={Initial Student Response}]
\ldots\ But there's a clever way to simplify it: $a+b+ab=(a+1)(b+1)-1$.
\ldots\ Let's denote $x=a+1$, $y=b+1$.
\ldots\ So the expression becomes: $n=xy-1$.
\ldots

\begin{tcolorbox}[casestep]
\caseannotation{Wrong reasoning step}\par
We want to \textbf{exclude perfect squares} (since in that case $x=y$)
\end{tcolorbox}

So first, let's list composite numbers $\leq101$, and then remove the perfect squares.
\ldots\
So we remove these 9 perfect squares from the list of 53 composite numbers.
\ldots\
So the total number of such $n$ is: $\boxed{44}$.
\end{tcolorbox}

\begin{tcolorbox}[title={\method-trained Student Response}]
\ldots\ Let $x=a+1$, $y=b+1$, so $x,y\geq2$ (since $a,b\geq1$), and $x\ne y$ (since $a\ne b$).
Then the expression becomes: $N=xy-1$.
\ldots\ Let's denote $M=N+1$.
\ldots

\begin{tcolorbox}[casestep]
\caseannotation{Wrong reasoning step}\par
So we need to count the number of integers $M\in[4,101]$ such that $M$ is composite and is \textbf{not a perfect square} (since perfect squares have only one distinct factorization into two equal integers)
\end{tcolorbox}

\begin{tcolorbox}[casestep]
\caseannotation{Reasoning revision}\par
Wait, not necessarily. For example, consider $M=16$.
It can be factored as $2\times8$, $4\times4$.
So even though it's a perfect square, it still can be factored into distinct integers.
So the perfect square condition is not sufficient to disqualify it.
\end{tcolorbox}

\ldots\ So the numbers that cannot be written as a product of two distinct integers $x,y\geq2$ are:
\begin{itemize}[leftmargin=*,nosep,topsep=2pt]
    \item Those composite numbers which are perfect squares of primes, i.e., numbers of the form $p^2$ where $p$ is prime
\end{itemize}
\ldots\ So, of the 74 composite numbers from 4 to 101, 4 of them cannot be written as a product of two distinct integers $\geq2$.
Therefore, the total number of integers $M\in[4,101]$ that can be written as $xy$ with $x,y\geq2$, $x\ne y$, is: $74-4=70$.
\ldots\ Final Answer: $\boxed{70}$.
\end{tcolorbox}

\smallskip
\end{minipage}
We find that both responses initially exclude all perfect squares, but the \method-trained student revises this rule after recognizing that $16=2\times8$ uses distinct factors.
It then excludes only squares of primes from the composite numbers and obtains the correct answer of 70, whereas the initial student retains the error and answers 44.

\par\medskip
\noindent\begin{minipage}{\linewidth}
\textbf{A competitive programming case}\label{app:code-revision-case}\par\smallskip
\begin{tcolorbox}[title={Question (TACO, Problem \#797C)}]
Petya received a gift: a string $s$ of length up to $10^5$ characters for his birthday.
He took two more empty strings $t$ and $u$ and decided to play a game.
This game has two possible moves:
\begin{itemize}[leftmargin=*,nosep,topsep=2pt]
    \item Extract the first character of $s$ and append $t$ with this character.
    \item Extract the last character of $t$ and append $u$ with this character.
\end{itemize}
Petya wants to make strings $s$ and $t$ empty and string $u$ lexicographically minimal.
\par Input: \texttt{acdb}\quad Output: \texttt{abdc}
\end{tcolorbox}

\begin{tcolorbox}[title={Initial Student Response}]
To solve this problem, we need to simulate the game Petya is playing.
The goal is to make the string \texttt{u} lexicographically minimal.
\ldots\ Use a stack (\texttt{t}) to simulate the intermediate string.

\begin{tcolorbox}[casestep]
\caseannotation{Wrong reasoning step}\par
At each step, check if the last character of the stack is less than or equal to the \textbf{next character} in \texttt{s}.
If so, pop it and add it to \texttt{u}.
\end{tcolorbox}

Otherwise, push the next character from \texttt{s} to the stack.
At the end, pop all remaining characters from the stack to \texttt{u}.
\ldots
\end{tcolorbox}

\begin{tcolorbox}[title={\method-trained Student Response}]
To solve this problem, we need to simulate the game that Petya is playing.
The goal is to have the final string \texttt{u} as lexicographically minimal as possible.
\ldots\ We can simulate the game by using a stack and a greedy approach.

\begin{tcolorbox}[casestep]
\caseannotation{Wrong reasoning step}\par
We push characters from \texttt{s} onto the stack, and when the top of the stack is less than or equal to the \textbf{next character} in \texttt{s}, we pop it to \texttt{u}.
\end{tcolorbox}

\ldots\ Wait, but this is not correct for the example ``\texttt{acdb}''.
\ldots\ \texttt{u} is \texttt{['a', 'c', 'b', 'd']} $\to$ but the correct answer is '\texttt{abdc}'.
So this algorithm is not correct.

\begin{tcolorbox}[casestep]
\caseannotation{Reasoning revision}\par
The correct solution is to compare the current top of the stack to the \textbf{minimum remaining character} in the remaining string.
\ldots\ We can precompute the minimum suffix for each position.
\end{tcolorbox}

\ldots\ Now, for each position \texttt{i}, after pushing \texttt{s[i]} to the stack, we can check if the top of the stack is less than or equal to \texttt{min\_suffix[i+1]}.
If so, it's safe to pop it to \texttt{u}.
\ldots
\begin{verbatim}
for i in range(n):
    t.append(s[i])
    while t and (i == n-1 or t[-1] <= min_suffix[i+1]):
        u.append(t.pop())
\end{verbatim}
\ldots
\end{tcolorbox}
\smallskip
\end{minipage}

We find that both students initially compare the stack top only with the next input character, overlooking smaller characters that appear later.
The \method-trained student detects this error on \texttt{acdb} and revises the comparison to use the smallest character in the remaining string.
Its final program returns the correct \texttt{abdc} instead of \texttt{acbd} and passes all validation tests.
\endgroup

\section{Training Algorithm}
\label{app:training}
\begin{algorithm}[H]
\caption{Algorithm of {\method}}
\label{alg:1}
\begin{algorithmic}[1]
\REQUIRE $\pi_{\theta}$, $\pi_{*}$, $\pi_\mathrm{ref}$, $x$, $\lambda$, $K$.
\ENSURE \method-trained student model $\pi_{\theta^+}$.
\STATE Initialize $\theta$ for student model $\pi_{\theta}$ and base reference policy $\pi_\mathrm{ref}$, respectively;
\STATE Select student segments $s$ from on-policy rollout $y$ for $x$ based on Eq.~(\ref{eq:segment-uncertainty});
    \WHILE{not converged yet}
        \FOR{$t=1,2,\ldots,N_{\mathrm{step}}$}
            \STATE Construct accepted revision units $p=(q,s,r)$ through teacher redrafting;
            \STATE Compute revision margins $m_{\pi_\theta}(p)$ based on Eq.~(\ref{eq:revision-margin});
            \STATE Compute $\mathcal{L}_s$ and $\mathcal{L}_t$ based on Eqs.~(\ref{eq:segment-loss}) and (\ref{eq:token-loss});
            \STATE Update $\theta$ by optimizing Eq.~(\ref{eq:seg-opd-objective});
        \ENDFOR
        \STATE Resample on-policy rollout $y$ and select $s$ based on Eqs.~(\ref{eq:pivot-entropy}) and (\ref{eq:segment-uncertainty});
    \ENDWHILE
\RETURN $\pi_{\theta^+}$;
\end{algorithmic}
\end{algorithm}

We formalize the training algorithm of \method in Algorithm~\ref{alg:1}.
In line 2, we first select up to $K$ student segments $s$ from the on-policy rollout $y$ using the uncertainty scores in Eq.~(\ref{eq:segment-uncertainty}).
From lines 3 to 11, we train the student by optimizing the joint objective in Eq.~(\ref{eq:seg-opd-objective}), where $N_{\mathrm{step}}$ denotes the number of training steps per iteration.
In detail, we construct accepted revision units $p=(q,s,r)$ through teacher redrafting (line 5), calculate their revision margins using Eq.~(\ref{eq:revision-margin}) (line 6), and compute the segment-wise preference and token-wise OPD losses using Eqs.~(\ref{eq:segment-loss}) and (\ref{eq:token-loss}) (line 7).
We then update the student parameters $\theta$ with these losses (line 8) and resample the on-policy rollout to select new student segments (line 10).
The teacher and reference policies remain fixed.
When no revision unit is accepted, we set $\mathcal{L}_s=0$ and retain all valid rollout tokens for token-wise OPD.
Next, in Appendix~\ref {app:complexity}, we analyze the training algorithm's time complexity.

\section{Complexity Analysis}\label{app:complexity}\label{app:complexity-analysis}

\begin{table}[th]
\centering
\caption{Relative training time on DeepMath-103K, normalized to OPD within student--teacher pair.}
\label{tab:training-complexity}
\label{tab:training-time}
\small
\renewcommand{\arraystretch}{1.1}
\resizebox{0.89\linewidth}{!}{%
\begin{tabular}{@{}llccc@{}}
\toprule
\textbf{Student} & \textbf{Teacher} & \textbf{OPD} & \textbf{OmniOPD} & \textbf{\method} \\
\midrule
DeepSeek-1.5B & JustRL-DeepSeek-1.5B & $1\times$ & $3.67\times$ & $1.89\times$ \\
Qwen3-4B & Qwen3-4B-Non-Thinking-RL-Math & $1\times$ & $3.89\times$& $1.93\times$ \\
Qwen3-8B & Qwen3-8B-DAPO & $1\times$ & $3.41\times$& $1.73\times$ \\
Qwen3-14B &  Qwen3-14B-Math-RL & $1\times$ & $3.39\times$ & $1.87\times$ \\
\bottomrule
\end{tabular}%
}
\end{table}

In this work, we propose \method, a segment-wise OPD method that enables students to learn reasoning revision from explicit teacher redrafting.
For an asymptotic complexity analysis, we treat model sizes, batch size, and the number of optimization passes as fixed and assume dense attention with KV-cached decoding.

Let $L$ be the total length of each problem prompt and its student rollout in an iteration.
Student generation, teacher scoring, and student optimization in OPD~\citep{agarwal2024policy} together cost $O(L(1+L))$ per iteration.
OmniOPD~\citep{zhou2026omniopd} samples $n$ teacher continuations of at most $c$ tokens at each of $K$ selected chunks, giving a conservative total bound of $O(L(1+L)+Kn(L+c)(1+L+c))$, including student optimization and reference-policy regularization.
For \method, let $D$ bound the number of new tokens in each full teacher draft before boundary alignment; generating up to $K$ drafts and learning from the revision pairs gives a total bound of $O(L(1+L)+K(L+D)(1+L+D))$.
These bounds include teacher-prefix prefill and reference-policy evaluation; boundary extraction adds only a linear scan of the generated text.
Here, $K$ denotes the per-rollout chunk or segment budget for the respective method; OmniOPD uses $K=10$, $n=10$, and $c=50$, whereas \method uses $K=1$ and redrafts only selected failed rollouts.
For fixed sampling budgets and $D=O(L)$, all three methods have quadratic upper bounds on sequence length, but differ in the amount of teacher generation and optimization work.

For practical cost, \method generates a complete teacher continuation from the selected prefix before extracting the leading sentence as $r$, which is detailed in Appendix~\ref{app:segment}.
Its generation cost therefore depends on the full draft rather than the retained redraft length; the subsequent extraction is negligible compared with model inference.
The additional generation work depends on how many rollouts are redrafted and on the full draft lengths produced from their selected prefixes.
Prefix processing and revision-pair optimization also contribute to training time.
OmniOPD instead generates up to $K \cdot n \cdot c=5{,}000$ teacher tokens across $K \cdot n=100$ continuations per rollout, and processes their conditioning prefixes.
We therefore report measured training time to quantify the practical cost of the additional revision supervision, including teacher generation and student optimization.
We compare with OPD as the token-wise baseline and with OmniOPD, which also generates teacher text for selected student chunks.

Table~\ref{tab:training-time} reports the measured training-time ratios for the mathematical reasoning runs under the implementation settings in Appendix~\ref{app:other}, normalized to OPD for each student--teacher pair.
\method timings include generating the full teacher drafts before extraction.
Across the four student–teacher pairs, Seg-OPD requires less measured training time than OmniOPD: $1.73$--$1.93\times$ the OPD training time, versus $3.39$--$3.89\times$ for OmniOPD.
Thus, \method's measured training time remains below twice that of OPD in these settings, including full teacher draft generation and preference learning.

\section{Discussions and Future Work}
In this work, we propose \method, a segment-wise OPD method that enables students to learn reasoning revision from explicit teacher redrafting.
We conduct extensive empirical analysis across diverse model architectures and downstream tasks, demonstrating our method's effectiveness in learning reasoning revision and improving student reasoning performance.
As with other post-training methods, \method may be misused to adapt LLMs for malicious purposes.
We encourage responsible use and appropriate safeguards to mitigate these risks.

Several promising directions emerge naturally for future research, such as extending \method to multi-teacher settings, including those with black-box teachers.
The segment-wise objective relies on preference supervision rather than teacher logits.
This would allow multiple teachers, including black-box models, to provide teacher redrafts for student segments, while a single white-box teacher supplies the logits required for token-wise OPD.
We believe our methodology can provide insights into exploring these directions simultaneously.

\end{document}